\documentclass[10pt,twocolumn,letterpaper]{article}

\usepackage{iccv}
\usepackage{times}
\usepackage{epsfig}
\usepackage{graphicx}
\usepackage{amsmath}
\usepackage{amssymb}

\usepackage[pagebackref=true,breaklinks=true,letterpaper=true,colorlinks,bookmarks=false]{hyperref}

\usepackage{color}
\usepackage{graphicx}
\usepackage{amsmath,amssymb} 
\usepackage{color}
\usepackage{xspace}
\usepackage{eso-pic}
\usepackage{epsfig}
\usepackage{amsmath}
\usepackage{amssymb}
\usepackage{bm}
\usepackage{subfigure}
\usepackage{algorithm}
\usepackage[update,prepend]{epstopdf} 
\usepackage{hyperref}
\usepackage{tabularx,ragged2e}

\usepackage{alltt}
\usepackage[square,comma,numbers,sort&compress]{natbib}
\usepackage{afterpage}
\RequirePackage[log]{snapshot}

 \graphicspath{{./images/}}%
 
\makeatletter
\DeclareRobustCommand\onedot{\futurelet\@let@token\@onedot}
\def\@onedot{\ifx\@let@token.\else.\null\fi\xspace}

\def\eg{\emph{e.g}\onedot} 
\def\ie{\emph{i.e}\onedot} 
 
\def\etc{\emph{etc}\onedot} 
 
\def\etal{\emph{et~al}\onedot}

\iccvfinalcopy %

\def\iccvPaperID{507} %

\ificcvfinal\fi
\begin{document}

\title{Flowing ConvNets for Human Pose Estimation in Videos}

\author{Tomas Pfister \\
Dept. of Engineering Science \\
University of Oxford \\
{\tt\small tp@robots.ox.ac.uk}
\and
James Charles \\
School of Computing \\
University of Leeds \\
{\tt\small j.charles@leeds.ac.uk}
\and
Andrew Zisserman \\
Dept. of Engineering Science \\
University of Oxford \\
{\tt\small az@robots.ox.ac.uk}
}

\maketitle

\begin{abstract}
The objective of this work is human pose estimation in videos, where
multiple frames are available.  We investigate a ConvNet architecture that is
able to benefit from temporal context by combining information across
the multiple frames using optical flow.

To this end we propose a network architecture with the following novelties: 
(i)~a deeper network than previously investigated for regressing heatmaps;
(ii)~spatial fusion layers that learn an implicit spatial model;
(iii)~optical flow is used to align heatmap predictions from neighbouring frames; and 
(iv)~a final parametric pooling layer which learns to combine the aligned heatmaps into a pooled confidence map.

We show that this architecture outperforms a number of others, including one that uses optical flow solely at the input layers, one that regresses joint coordinates directly, and one that predicts heatmaps without spatial fusion.

The new architecture outperforms the state of the art by a large margin on three video pose estimation datasets, including the very challenging Poses in the Wild dataset, and outperforms other deep methods that don't use a graphical model on the single-image FLIC benchmark (and also~\cite{Chen14,Tompson14a} in the high precision region).
\end{abstract}

\section{Introduction}

Despite a long history of research, human pose estimation in videos remains a very challenging task in computer vision.  
Compared to still image pose estimation, the temporal component of videos provides an additional (and important) cue for recognition, as strong dependencies of pose positions exist between temporally close video frames.

\begin{figure*}[t]
\centering
\includegraphics[width=\linewidth]{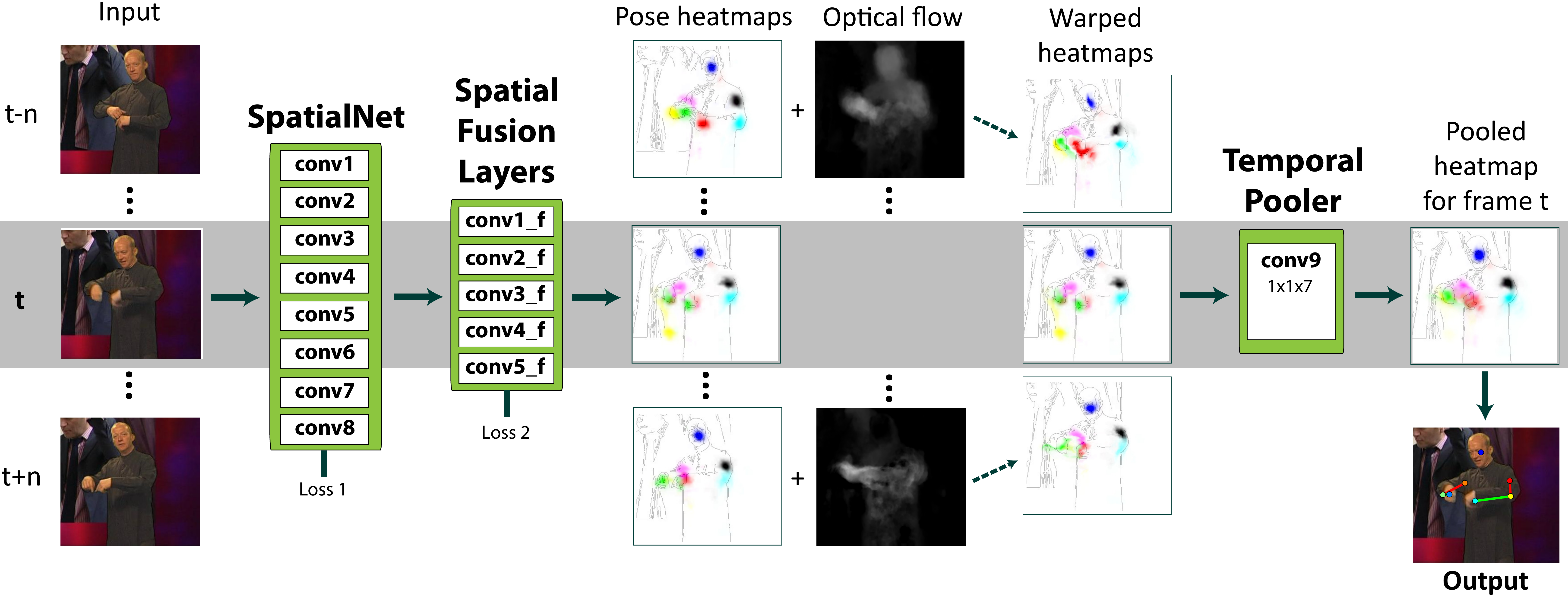}
\vspace{-0.4cm}
\caption{{\bf Deep expert pooling architecture for pose estimation.} 
The network takes as an input all RGB frames within a $n$-frame neighbourhood of the current frame $t$.
The fully convolutional network (consisting of a heatmap net with an implicit spatial model) predicts a confidence heatmap for each body joint in these frames (shown here with a different colour per joint).
These heatmaps are then temporally \emph{warped} to the current frame $t$ using optical flow.
The warped heatmaps (from multiple frames) are then \emph{pooled} with another convolutional layer (the temporal pooler), which learns how to weigh the warped heatmaps from nearby frames.
The final body joints are selected as the maximum of the pooled heatmap (illustrated here with a skeleton overlaid on top of the person). 
}
\label{fig:teaser}
\end{figure*}

In this work we propose a new approach for using optical flow for part localisation in deep Convolutional Networks (ConvNets), and demonstrate its performance for human pose estimation in videos.  
The key insight is that, since for localisation the prediction targets are positions in the image space (\eg $(x,y)$ coordinates of joints), one can use dense optical flow vectors to \emph{warp predicted positions} onto a target image.  
In particular, we show that when regressing a \emph{heatmap} of positions (in our application for human joints), the heatmaps from neighbouring frames can be warped and aligned using optical flow, effectively propagating position confidences temporally, as illustrated in Fig~\ref{fig:teaser}.

We also propose a deeper architecture that has additional convolutional layers beyond the initial heatmaps to enable learning an \emph{implicit} spatial model of human layout. 
These layers are able to learn dependencies between human body parts. 
We show that these `spatial fusion' layers remove pose estimation failures that are kinematically impossible.

\vspace{-0.4cm}\paragraph{Related work.}
\label{sec:related}
Traditional methods for pose estimation have often used pictorial structure models~\cite{Buehler11,Eichner12,Felzenszwalb05,Sapp11,Yang13a}, which optimise a configuration of parts as a function of local image evidence for a part, and a prior for the relative positions of parts in the human kinematic chain. 
An alternative approach uses poselets~\cite{Bourdev09,Gkioxari14}.  
More recent work has tackled pose estimation holistically: initially with Random Forests on depth data~\cite{Shotton13,Girshick11,Taylor12,Sun12} and RGB~\cite{Charles13a,Charles14,Pfister14a}, and most recently with convolutional neural networks.

The power of ConvNets has been demonstrated in a wide variety of vision tasks -- object classification and detection~\cite{Krizhevsky12,Zeiler13,Sermanet14,Girshick14}, face recognition~\cite{Taigman14}, text recognition~\cite{Goodfellow13,Jaderberg14a}, video action
recognition~\cite{Karpathy14,Simonyan14} and many more~\cite{Donahue13,Sharif14,Osadchy07}.

For pose estimation, there were early examples of using ConvNets for pose comparisons~\cite{Taylor10}. 
More recently, \cite{Toshev13} used an AlexNet-like ConvNet to directly regress \emph{joint coordinates}, with a cascade of ConvNet regressors to improve accuracy over a single pose regressor network. 
Chen and Yuille~\cite{Chen14} combine a parts-based model with ConvNets (by using a ConvNet to learn conditional probabilities for the presence of parts and their spatial relationship with image patches).  
In a series of papers, Tompson, Jain~\etal developed ConvNet architectures to directly regress {\em heatmaps} for each joint, with subsequent layers to add an Markov Random Field (MRF)-based spatial model~\cite{Tompson14,Jain14}, and a pose refinement model~\cite{Tompson14a} (based on a Siamese network with shared weights) upon a rougher pose estimator ConvNet.

Temporal information in videos was initially used with ConvNets for
action recognition~\cite{Simonyan14}, where optical flow was used as
an input \emph{motion feature} to the network.  Following this work,
\cite{Jain14a,Pfister14a} investigated the use of temporal information
for pose estimation in a similar manner, by inputting flow or RGB from
multiple nearby frames into the network, and predicting joint
positions in the current frame.

However, pose estimation differs from action recognition in a key
respect which warrants a different approach to using optical flow: in
action recognition the prediction target is a class label, whereas in
pose estimation the target is a set of $(x,y)$ positions onto the
image.  Since the targets are positions in the image space, one can
use dense optical flow vectors not only as an input feature but also
to \emph{warp predicted positions} in the image, as done
in~\cite{Charles14} for random forest estimators.  
To this end, our work explicitly predicts joint
positions for \emph{all} neighbouring frames, and temporally aligns
them to frame $t$ by warping them backwards or forwards in time using
tracks from dense optical flow.  This effectively reinforces the
confidence in frame $t$ with a strong set of `expert opinions' 
(with corresponding confidences)
from
neighbouring frames, from which joint positions can be more precisely
estimated.  Unlike~\cite{Charles14} who average the expert opinions,
we learn the expert pooling weights with backpropagation in an
end-to-end ConvNet.

Our ConvNet outperforms the state of the art on three challenging video pose estimation
datasets (BBC Pose, ChaLearn and Poses in the Wild) -- the heatmap regressor
alone surpasses the state of the art on these datasets, and the pooling from
neighbouring frames using optical flow gives a further significant boost.
We have released the models and code at \url{http://www.robots.ox.ac.uk/~vgg/software/cnn_heatmap}.

\section{Temporal Pose Estimation Networks}
\label{sec:method}

Fig~\ref{fig:teaser} shows an overview of the ConvNet architecture.
Given a set of input frames within a temporal neighbourhood of $n$ frames from a frame $t$, a spatial ConvNet regresses joint confidence maps (`heatmaps') for each input frame separately.
These heatmaps are then individually \emph{warped} to frame $t$ using dense optical flow.
The warped heatmaps (which are effectively `expert opinions' about joint positions from the past and future) are then pooled into a single heatmap for each joint, from which the pose is estimated as the maximum.

We next discuss the architecture of the ConvNets in detail.
This is followed by a description of how optical flow is used to warp and pool the output from the Spatial ConvNet.

\subsection{Spatial ConvNet}

\begin{figure}[t]
\centering
\includegraphics[width=\linewidth]{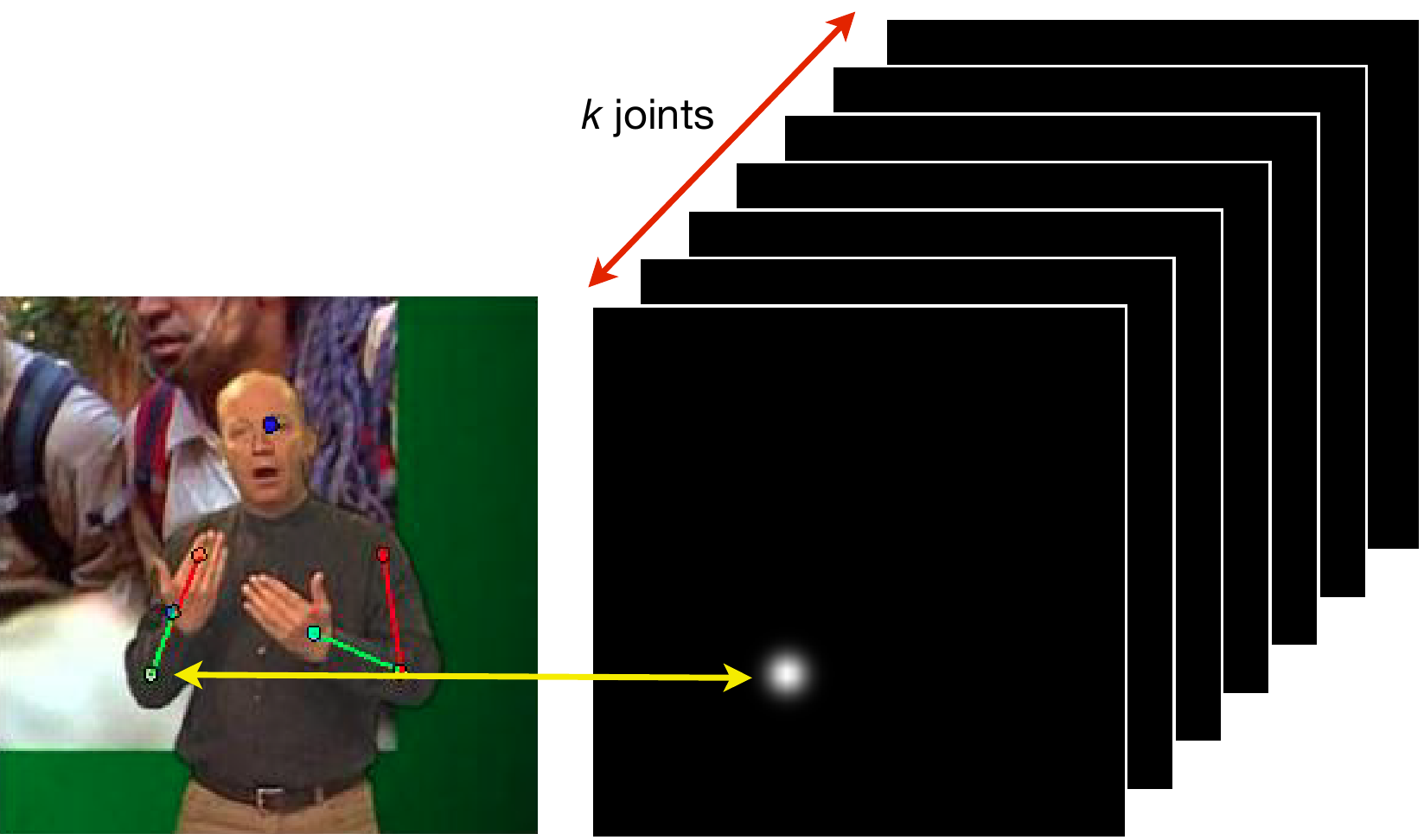}
\caption{{\bf Regression target for learning the Spatial ConvNet.} 
The learning target for the convolutional network is (for each of $k$ joints) a heatmap with a synthesised Gaussian with a fixed variance centred at the ground truth joint position.
The loss is $l_2$ between this target and the output of the last convolutional layer.
}
\label{fig:target}
\end{figure}

The network is trained to regress the location of the human joint positions.  
However, instead of regressing the joint $(x,y)$ positions directly~\cite{Toshev13,Pfister14a}, we regress a \emph{heatmap} of the joint positions, separately for each joint in an input image. 
This heatmap (the output of last convolutional layer, conv8) is a fixed-size $i \times j \times k$-dimensional cube (here $64 \times 64 \times 7$ for $k=7$
upper-body joints).  
At training time, the ground truth label are heatmaps synthesised for each joint separately by placing a Gaussian with fixed variance at the ground truth joint position (see Fig~\ref{fig:target}).  
We then use an $l_2$ loss, which penalises the squared pixel-wise differences between the predicted heatmap and the synthesised ground truth heatmap.

We denote the training example as $(\mathbf X, \mathbf y)$, where $\mathbf y$ stands for the coordinates of the $k$ joints in the image $\mathbf X$. 
Given training data $N=\{\mathbf X, \mathbf y\}$ and the ConvNet regressor $\phi$ (output from conv8), the training objective becomes the task of estimating the network weights $\lambda$:
\begin{equation}
\arg \min_{\lambda} \sum_{(\mathbf X, \mathbf y) \in N} \sum_{i, j, k} \| G_{i,j,k}(\mathbf y_k) - \phi_{i,j,k}(\mathbf X, \lambda) \|^2
\end{equation}
where $G_{i,j,k}(\mathbf y_i) = \frac{1}{{2\pi \sigma^2 }}e^{{{ -[ (y_k^1 - i)^2 } + (y_k^2 - j)^2] / {2\sigma ^2 }}}$ is a Gaussian centred at joint $y_k$ with fixed $\sigma$.

\begin{figure}[t]
\centering
\includegraphics[width=0.9\linewidth]{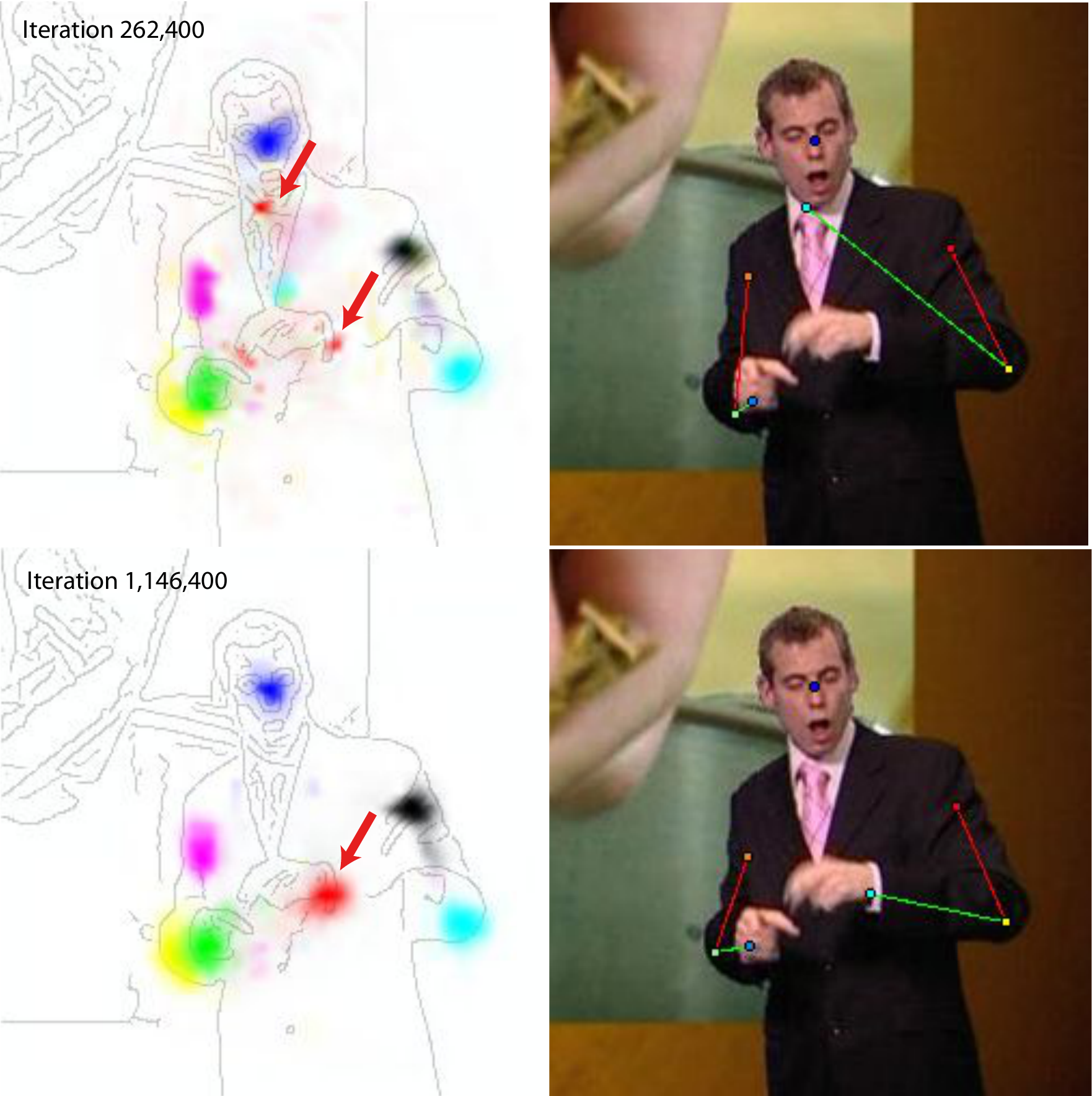}
\caption{{\bf Multiple peaks are possible with the Spatial ConvNet.} 
Early on in training (top), multiple locations may fire for a given
joint.  These are then suppressed as training proceeds (bottom). The
arrows identify two modes for the wrist; one correct, one erroneous. As the
training proceeds the erroneous one is diminished.  }
\label{fig:multipeak}
\end{figure}

\begin{figure}[t]
\centering
\includegraphics[width=\linewidth]{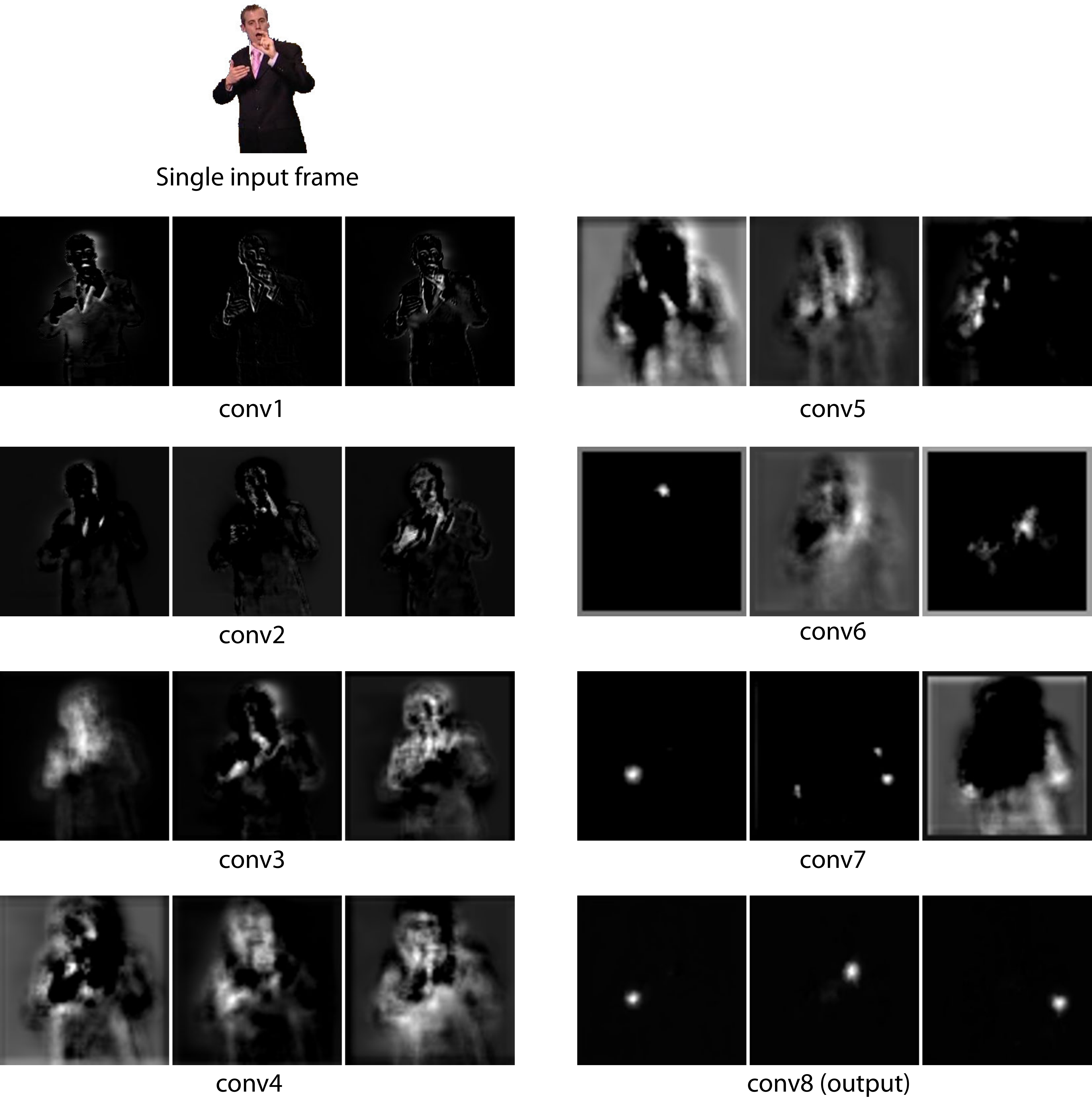}
\caption{{\bf Sample activations for convolutional layers.} 
Neuron activations are shown for three randomly selected channels for each convolutional layer (resized here to the same size), with the input (pre-segmented for visualisation purposes) shown above. 
Low down in the net, neurons are activated at edges in the image (\eg conv1 and conv2); higher up, they start responding more clearly to body parts (conv6 onwards).
The outputs in conv8 are shown for the right elbow, left shoulder and left elbow.
}
\label{fig:activations}
\end{figure}

\vspace{-0.4cm}\paragraph{Discussion.}
As noted by~\cite{Tompson14}, regressing coordinates
directly is a highly
non-linear and more difficult to learn mapping, which we also confirm
here (Sect~\ref{sec:experiments}).
The benefits of regressing a heatmap rather than $(x,y)$ coordinates
are twofold: first, one can understand failures and visualise the
`thinking process' of the network (see Figs~\ref{fig:multipeak}
and~\ref{fig:activations}); second, since by design, the output of the
network can be multi-modal, \ie allowed to have confidence at multiple
spatial locations, learning becomes easier: early on in training (as
shown in Fig~\ref{fig:multipeak}), multiple locations may fire for a
given joint; the incorrect ones are then slowly
suppressed as training proceeds.  In contrast, if the output were only
the wrist $(x,y)$ coordinate, the net would only have a lower loss if
it gets its prediction right (even if it was `growing confidence' in
the correct position).

\vspace{-0.4cm}\paragraph{Architecture.} 
The network architecture is shown in Fig~\ref{fig:teaser}, and sample
activations for the layers are shown in Fig~\ref{fig:activations}.  To
maximise the spatial resolution of the heatmap we make
two important design choices: 
(i)~minimal pooling is used (only two $2 \times 2$ max-pooling layers), and 
(ii)~all strides are unity (so that the resolution is not reduced).  
All layers are followed by ReLUs except conv9 (the pooling layer).
In contrast to AlexNet~\cite{Krizhevsky12},
our network is fully convolutional (no fully-connected layers) with
the fully-connected layers of~\cite{Krizhevsky12} replaced by $1
\times 1$ convolutions.  
In contrast to both AlexNet
and~\cite{Tompson14a}, our network is deeper, does not use local contrast normalisation (as we did not find this beneficial), and utilises less
max-pooling.

\begin{figure*}[t]
\centering
\includegraphics[width=\linewidth]{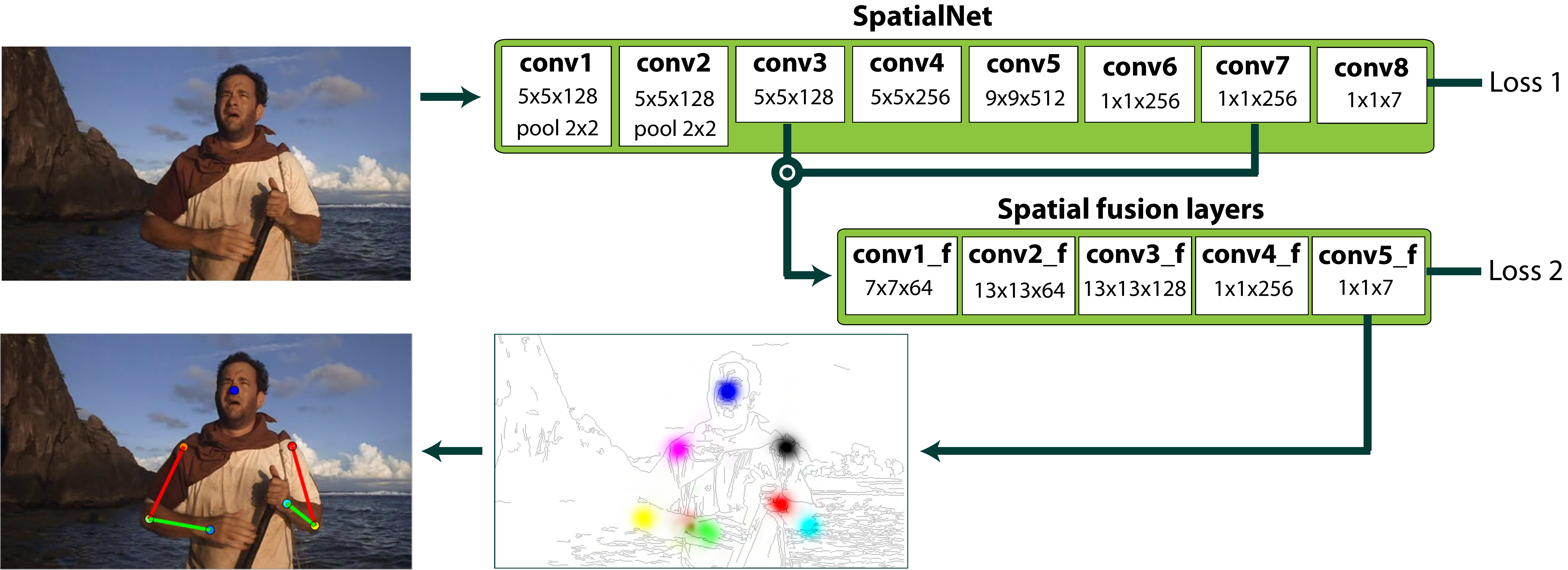}
\vspace{-0.2cm}
\caption{{\bf Spatial fusion layers.} 
The fusion layers learn to encode dependencies between human body parts locations, learning an implicit spatial model.
}
\label{fig:fusion}
\end{figure*}

\subsection{Spatial fusion layers}
Vanilla heatmap pose nets do not learn spatial dependencies of joints, and thus often predict kinematically impossible poses (see Fig~\ref{fig:failures}). 
To address this, we add what we term `spatial fusion layers' to the network.
These spatial fusion layers (normal convolutional layers) take as an input pre-heatmap activations (conv7), and learn dependencies between the human body parts locations represented by these activations.
In detail, these layers take as an input a concatenation of conv7 and conv3 (a skip layer), and feed these through five more convolutional layers with ReLUs (see Fig~\ref{fig:fusion}).
Large kernels are used to inflate the receptive field of the network.
We attach a separate loss layer to the end of this network and backpropagate through the whole network.

\subsection{Optical flow for pose estimation}

Given the heatmaps from the Spatial ConvNet from multiple frames, the heatmaps are reinforced with optical flow.
This is done in three steps:
(1)~the confidences from nearby frames are aligned to the current frame using dense optical flow;
(2)~these confidences are then \emph{pooled} into a composite confidence map using an additional convolutional layer; and
(3)~the final upper body pose estimate for a frame is then simply the positions of maximum confidence from the composite map.
Below we discuss the first two steps.

\def \imgscale {0.5}
\def \imgscaletwo {0.135}
\def \imgscalethree {0.32}
\begin{figure*}[t]
\begin{center}
\begin{tabular}{c*{3}{@{\hspace{1pt}}c}}
\raisebox{3mm}{\includegraphics[scale=\imgscale]{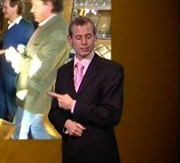}} &
\hspace{0.3cm} \raisebox{3mm}{\includegraphics[scale=\imgscalethree]{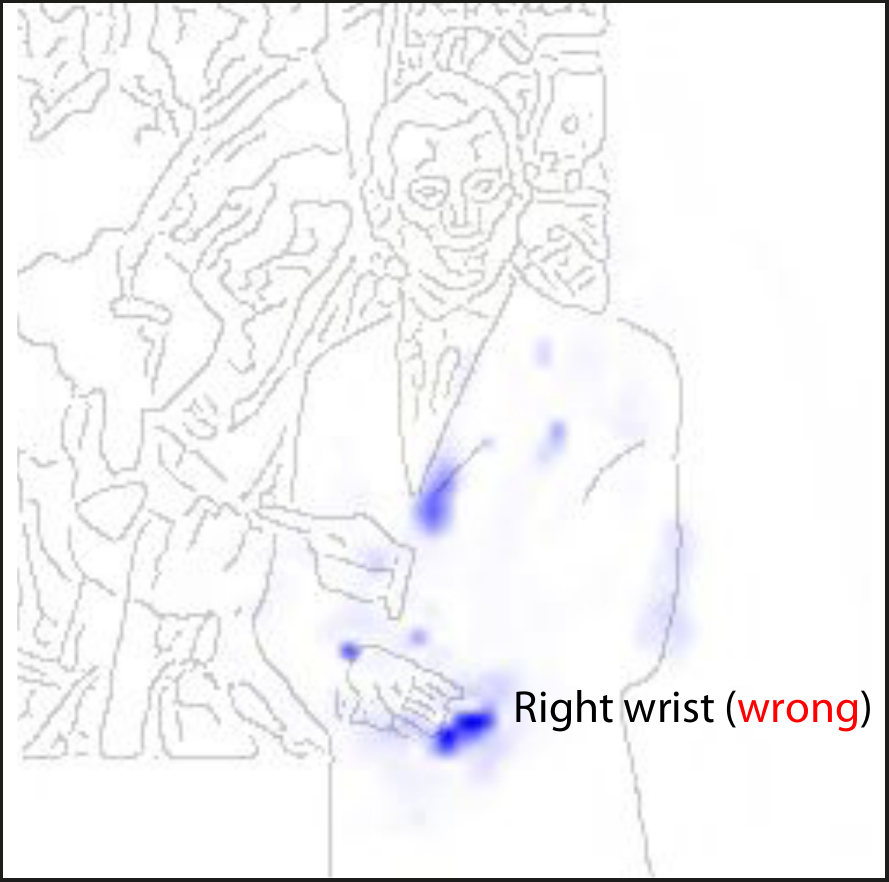}} \hspace{2mm}&
\includegraphics[scale=\imgscaletwo]{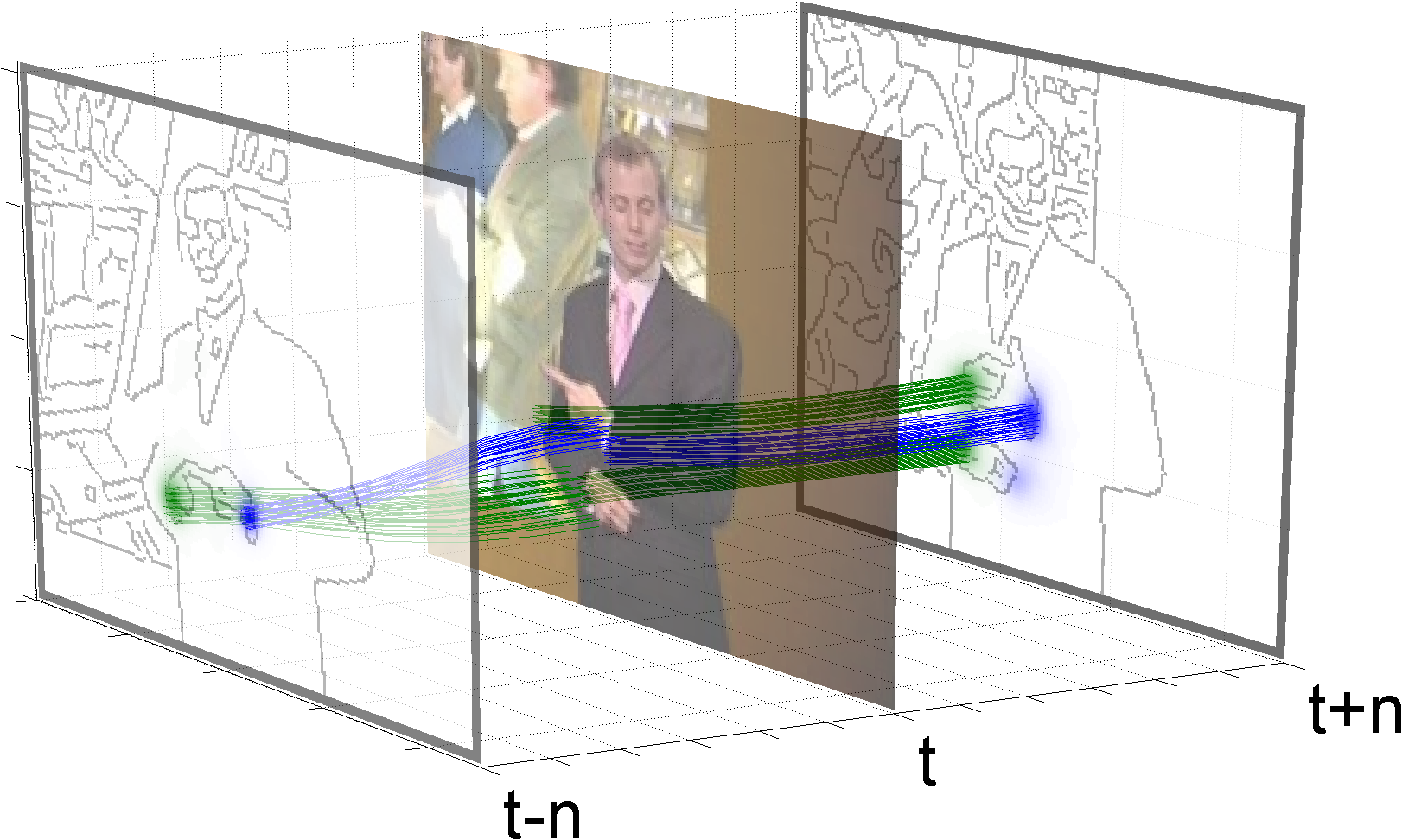} &
\hspace{0.3cm}\raisebox{3mm}{\includegraphics[scale=\imgscalethree]{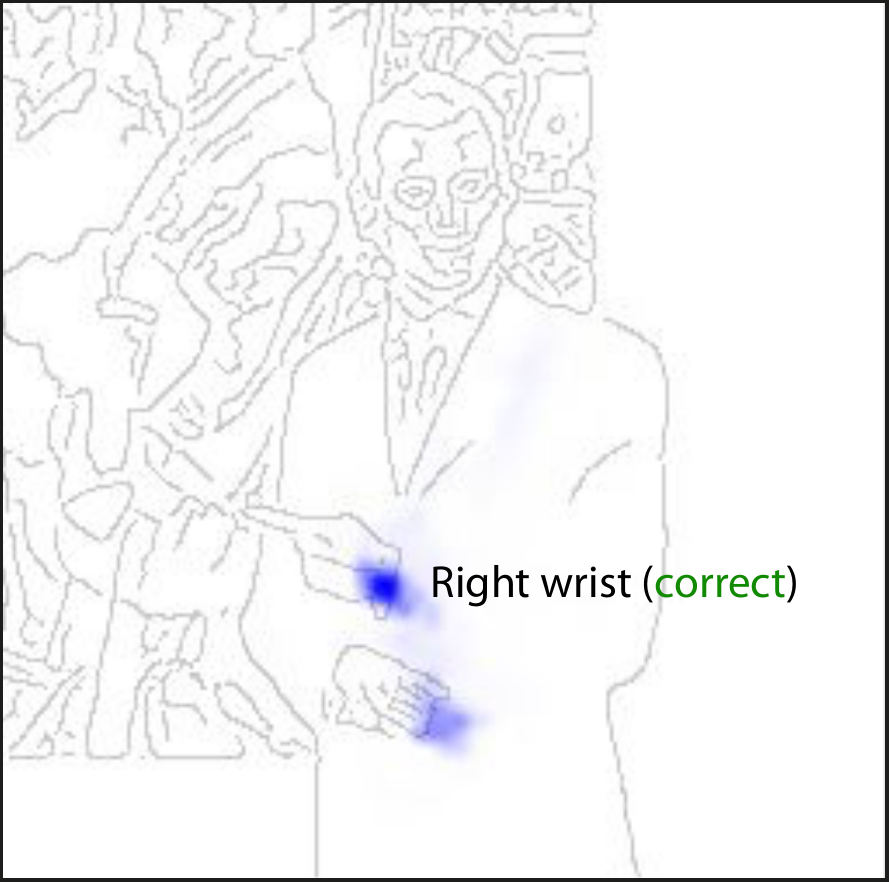}} \\
{\small (a) Frame $t$} & {\small(b) Confidence at $t$} & {\small(c) Warping} & {\small(d) Pooled at $t$}
\end{tabular}
\end{center}
\caption{{\bf Warping neighbouring heatmaps for improving pose estimates.} 
(a)~RGB input at frame $t$. 
(b)~Heatmap at frame $t$ for right hand in the image. 
(c)~Heatmaps from frames $(t-n)$ and $(t+n)$ warped to frame $t$ using tracks from optical flow (green \& blue lines). 
(d)~Pooled confidence map with corrected modes.}
\label{fig:warping}
\end{figure*}

\vspace{-0.4cm}\paragraph{Step 1: Warping confidence maps with optical flow.}
For a given frame~$t$, pixel-wise temporal tracks are computed from all neighbouring frames within $n$ frames from ($(t-n)$ to $(t+n)$) to frame~$t$ using dense optical flow~\cite{Weinzaepfel13}. 
These optical flow tracks are used to warp confidence values in neighbouring confidence maps to align them to frame~$t$ by effectively shifting confidences along the tracks~\cite{Charles14}.
Example tracks and the warping of wrist confidence values are shown in Fig~\ref{fig:warping}.

\vspace{-0.4cm}\paragraph{Step 2: Pooling the confidence maps.}
The output of Step~1 is a set of confidence maps that are warped to frame $t$.
From these `expert opinions' about the joint positions, the task is first to select a confidence for each pixel for each joint, and then to select one position for each joint.
One solution would be to simply average the warped confidence maps.
However, not all experts should be treated equally: intuitively, frames further away (thus with more space for optical flow errors) should be given lower weight.

To this end we learn a parametric cross-channel pooling layer that takes as an input a set of warped heatmaps for a given joint, and as an output predicts a single `composite heatmap'.
The input to this layer is a $i \times j \times t$ heatmap volume, where $t$ is the number of warped heatmaps (\eg 31 for a neighbourhood of $n=15$).
As the pooling layer, we train a $1 \times 1$ kernel size convolutional layer for each joint.
This is equivalent to cross-channel weighted sum-pooling, where we learn a single weight for each input channel (which correspond to the warped heatmaps).
In total, we therefore learn $t \times k$ weights (for $k$ joints).

\section{Implementation Details}
\label{sec:impdetails}

\paragraph{Training.} 
The input frames are rescaled to height 256. A $248 \times 248$ sub-image (of the $N \times 256$ input image) is randomly cropped, randomly horizontally flipped, randomly rotated between $-40^{\circ}$ and $-40^{\circ}$, and resized to $256 \times 256$.
Momentum is set to 0.95.
The variance of the Gaussian is set to $\sigma = 1.5$ with an output heatmap size of $64 \times 64$.
A temporal neighbourhood of $n=15$ is input into the parametric pooling layer.
The learning rate is set to $10^{-8}$, and decreased to $10^{-9}$ at 80K iterations, to $10^{-10}$ after 100K iterations and stopped at 120K iterations. 
We use Caffe~\cite{Jia13}.

\vspace{-0.4cm}\paragraph{Training time.} 
Training was performed on four NVIDIA GTX Titan GPUs using a modified version of the Caffe framework~\cite{Jia13} with multi-GPU support. 
Training SpatialNet on FLIC took 3 days \& SpatialNet Fusion 7 days.

\vspace{-0.4cm}\paragraph{Optical flow.}
Optical flow is computed using FastDeepFlow~\cite{Weinzaepfel13} with Middlebury parameters.

\section{Datasets}
\label{sec:data}

Experiments in this work are conducted on four large video pose estimation datasets, two from signed TV broadcasts, one of Italian gestures, and the third of Hollywood movies.
An overview of the datasets is given in Table~\ref{table:data}.
The first three datasets are available at \url{http://www.robots.ox.ac.uk/~vgg/data/pose}.

\begin{table}
\footnotesize
\centering
\begin{tabular}{| l | r | r | r | r |}
\hline
 & {\bf BBC} 	& {\bf Ext. BBC} 			& {\bf ChaLearn} 		& {\bf PiW}  \\ \hline
Train frames 	& 1.5M	& 7M    			& 1M				& 4.5K (FLIC)		\\ \hline
Test frames	& 1,000  & 1,000 	& 3,200 			& 830  \\ \hline
Train labels 	& \cite{Buehler11}  	& \cite{Buehler11,Charles13a}    & Kinect				& Manual		\\ \hline
Test labels 	& Manual 	& Manual    	& Kinect			& Manual		\\ \hline
Train videos 	& 13		& 85 				& 393				& -		\\ \hline
Val videos 	& 2		& 2 	   			& 287				& -		\\ \hline
Test videos 	& 5  		& 5 		    		& 275				& 30		\\ \hline
\end{tabular}
\caption{{\bf Dataset overview, including train/val/test splits.} 
}
\label{table:data}
\end{table}

\vspace{-0.4cm}\paragraph{BBC Pose dataset.}
This dataset~\cite{Charles13a} consists of 20 videos (each 0.5h--1.5h in length) recorded from the BBC with an overlaid sign language interpreter.
Each frame has been assigned pose estimates using the semi-automatic but reliable pose estimator of Buehler~\etal~\cite{Buehler11} (used as training labels).
1,000 frames in the dataset have been manually annotated with upper-body pose (used as testing labels).

\vspace{-0.4cm}\paragraph{Extended BBC Pose dataset.}
This dataset~\cite{Pfister14a} contains 72 additional training videos which, combined with the original BBC TV dataset, yields in total 85 training videos.
The frames of these new videos have been assigned poses using the automatic tracker of Charles~\etal~\cite{Charles13a}.
The output of this tracker is noisier than the semi-automatic tracker of Buehler~\etal, which results in partially noisy annotations.

\vspace{-0.4cm}\paragraph{ChaLearn dataset.}
The ChaLearn 2013 Multi-modal gesture dataset~\cite{Escalera13} contains 23 hours of Kinect data of 27 people. 
The data includes RGB, depth, foreground segmentations and full body skeletons.
In this dataset, both the training and testing labels are noisy (from Kinect).
The large variation in clothing across videos poses a challenging task for pose estimation methods.

\vspace{-0.4cm}\paragraph{Poses in the Wild (PiW) and FLIC datasets.}
The Poses in the Wild dataset~\cite{Cherian14} contains 30 sequences (total 830 frames) extracted from Hollywood movies.
The frames are annotated with upper-body poses. 
It contains realistic poses in indoor and outdoor scenes, with background clutter, severe camera motion and occlusions.
For training, we follow~\cite{Cherian14} and use all the images annotated with upper-body parts (about 4.5K) in the FLIC dataset~\cite{Sapp13}.

\section{Experiments}
\label{sec:experiments}

We first describe the evaluation protocol, then present comparisons to alternative network architectures, and finally give a comparison to state of the art.
A demo video is online at \url{https://youtu.be/yRLOid4XEJY}.

\subsection{Evaluation protocol and details}
\paragraph{Evaluation protocol.} 
In all pose estimation experiments we compare the estimated joints against frames with manual ground truth (except ChaLearn, where we compare against output from Kinect).
We present results as graphs that plot accuracy vs distance from ground truth in pixels, where a joint is deemed correctly located if it is within a set distance of $d$ pixels from a marked joint centre in ground truth.

\vspace{-0.4cm}\paragraph{Experimental details.}
All frames of the training videos are used for training (with each frame randomly augmented as detailed above). 
The frames are randomly shuffled prior to training to present maximally varying input data to the network.
The hyperparameters (early stopping, variance $\sigma$ \etc) are estimated using the validation set.  

\vspace{-0.4cm}\paragraph{Baseline method.}
As a baseline method we include a CoordinateNet (described in~\cite{Pfister15}).  
This is a network with similar architecture to~\cite{Sermanet14}, but trained for regressing the joint positions directly (instead of a heatmap)~\cite{Pfister14a}.

\vspace{-0.4cm}\paragraph{Computation time.}
Our method is real-time (50fps on 1 GPU without optical flow, 5fps with optical flow).

\subsection{Component evaluation}

\begin{figure}
\vspace{-0.2cm}
\centering
\includegraphics[width=\linewidth]{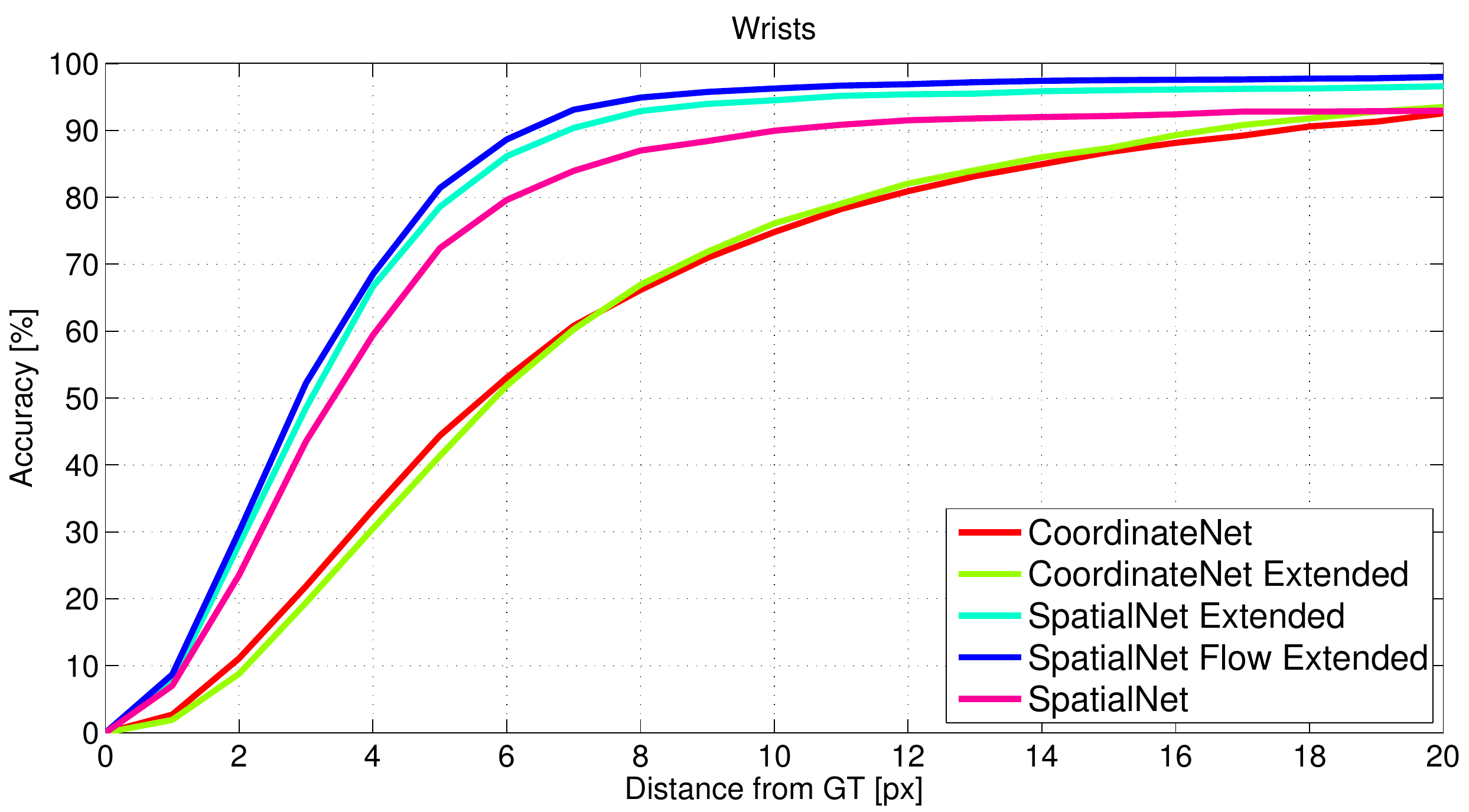}
\vspace{-0.4cm}
\caption{{\bf Comparison of the performance of our nets for wrists on BBC Pose.} 
Plots show accuracy as the allowed distance from manual ground truth is
increased. CoordinateNet is the network in~\cite{Pfister15}; SpatialNet is the heatmap network; and
SpatialNet Flow is the heatmap network with the parametric pooling
layer.  `Extended' indicates that the network is trained on Extended
BBC Pose instead of BBC Pose.  We observe a significant gain for the
SpatialNet from using the additional training data in the Extended BBC dataset (automatically labelled -- see
Sect~\ref{sec:data}) training data, and a further boost from using
optical flow information (and selecting the warping weights with the
parametric pooling layer).  }
\label{fig:ournets}
\end{figure}

For these experiments the SpatialNet and baseline are trained and 
tested on the BBC Pose and Extended BBC Pose datasets.
Fig~\ref{fig:ournets} shows the results for wrists

With the SpatialNet, we observe a significant boost in performance
(an additional 6.6\%, from 79.6\% to 86.1\% at $d=6$) when training on
the larger Extended BBC dataset compared to the BBC Pose dataset.
As noted in Sect~\ref{sec:data},
this larger dataset is somewhat noisy. In contrast,
the CoordinateNet is unable to make effective use of this additional
noisy training data.  We believe this is because its target (joint coordinates) does not allow for multi-modal output, which
makes learning from noisy annotation challenging.

\begin{figure}
\centering
\includegraphics[width=0.82\linewidth]{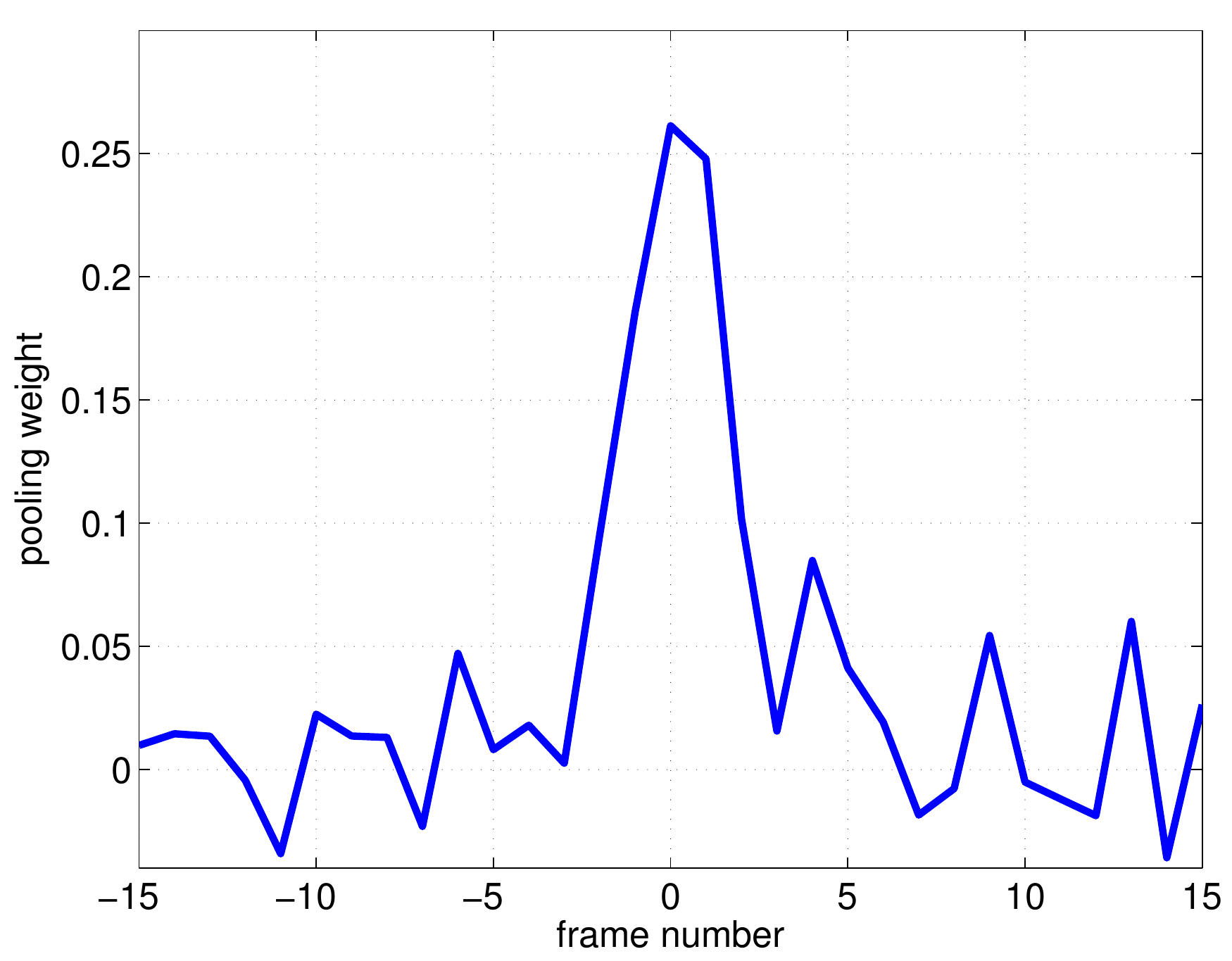}
\caption{{\bf Learnt pooling weights for BBC Pose with $n=15$.} 
Weights shown for the right wrist.
The centre frame receives highest weight.
The jitter in weights is due to errors in optical flow computation (caused by the moving background in the video) -- the errors become larger further away from the central frame (hence low or even negative weights far away).
}
\label{fig:weights}
\end{figure}

We observe a further boost in performance from using optical flow to warp heatmaps from neighbouring frames (an improvement of 2.6\%, from 86.1\% to 88.7\% at $d=6$).
Fig~\ref{fig:weights} shows the automatically learnt pooling weights. 
We see that for this dataset, as expected, the network learns to weigh frames temporally close to the current frame higher (because they contain less errors in optical flow). 

\begin{figure}
\centering
\vspace{-0.3cm}
\includegraphics[width=0.82\linewidth]{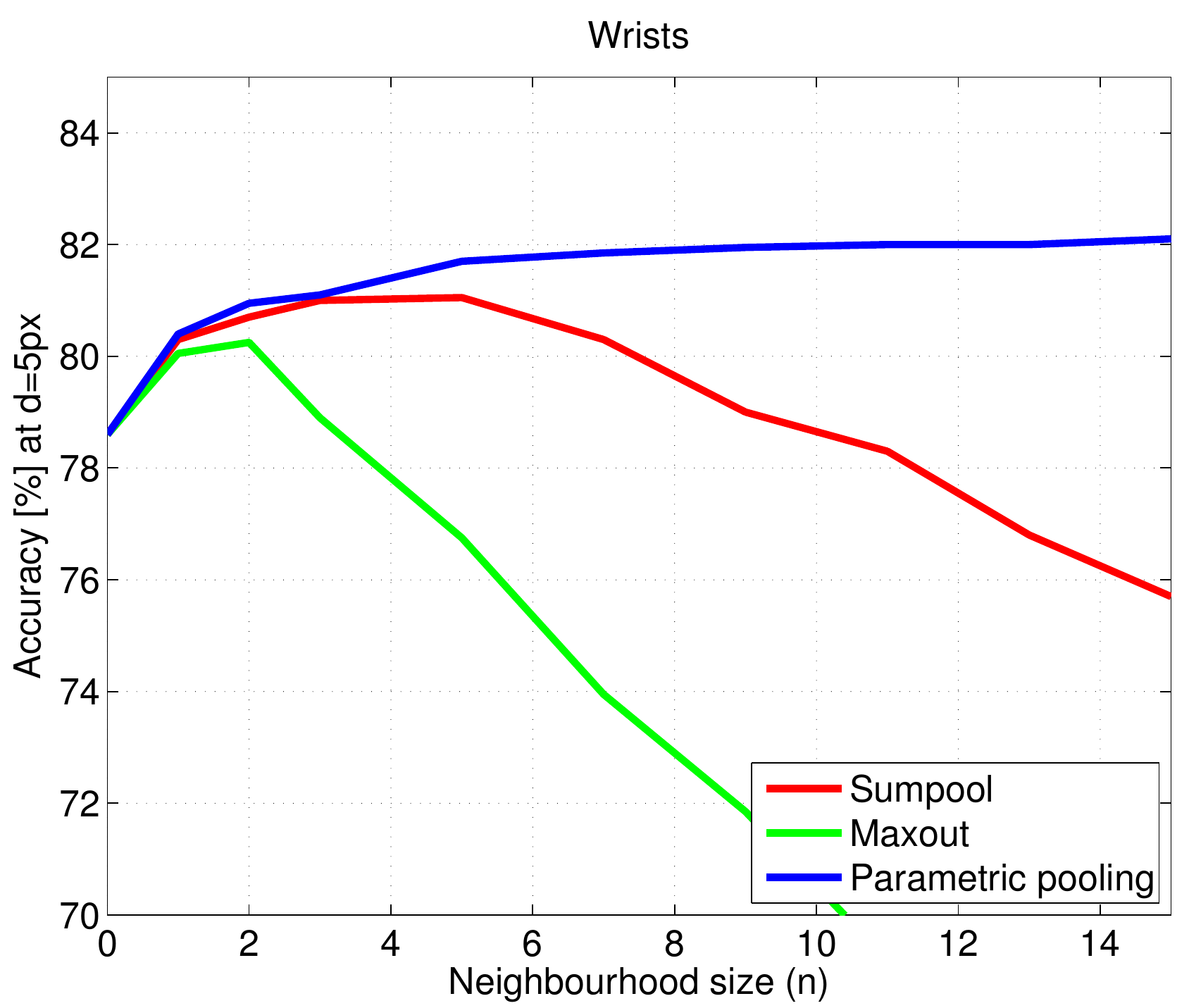}
\caption{{\bf Comparison of pooling types.}
Results are shown for wrists in BBC Pose at threshold $d=5$px.
Parametric pooling (learnt cross-channel pooling weights) performs best.}
\label{fig:optimisen}
\end{figure}

Fig~\ref{fig:optimisen} shows a comparison of different pooling types (for cross-channel pooling).
We compare learning a parametric pooling function to sum-pooling and
to max-pooling (maxout~\cite{Goodfellow13a}) across channels.  As
expected, parametric pooling performs best, and improves as the
neighbourhood $n$ increases.  In contrast, results with both
sum-pooling and max-pooling deteriorate as the neighbourhood size is
increased further, as they are not able to down-weigh predictions that
are further away in time (and thus more prone to errors in optical
flow).  
As expected, this effect is particularly noticeable for
max-pooling.

\begin{figure*}
\centering
\begin{minipage}[b]{0.44\textwidth}
\includegraphics[width=\linewidth]{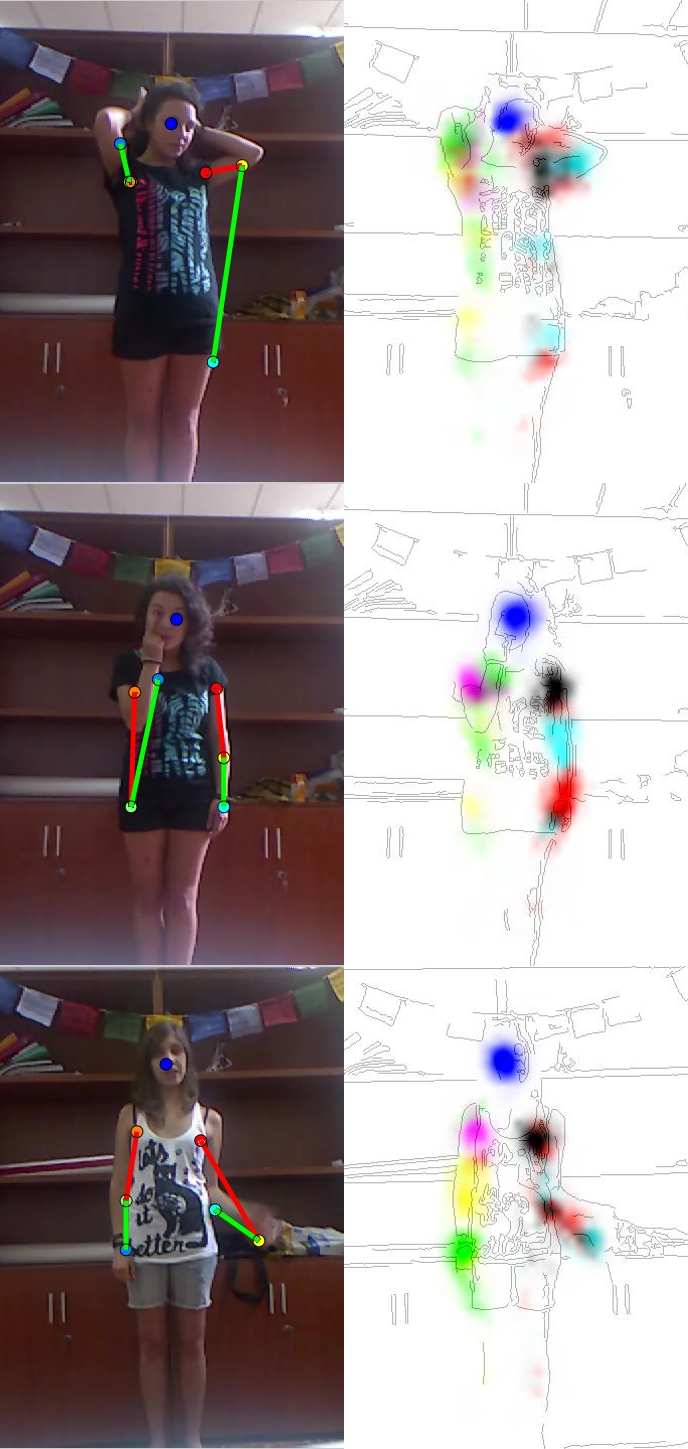}
\end{minipage}
\hfill
\begin{minipage}[b]{0.44\textwidth}
\includegraphics[width=\linewidth]{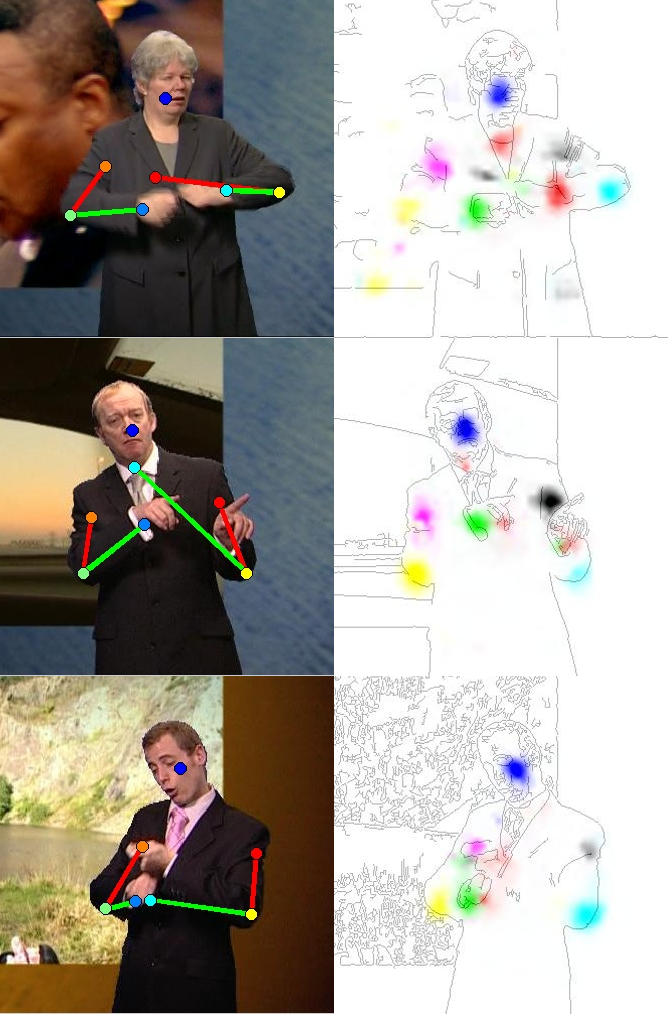}
\end{minipage}
\vspace{0.2cm}
\caption{{\bf Failures cases.} 
As shown, failure cases contain multiple modes for the same joint in the heatmap (and the wrong mode has been selected).
Adding spatial fusion layers (an implicit spatial model) resolves these failures.
}
\label{fig:failures}
\end{figure*}

\vspace{-0.4cm}\paragraph{Failure modes.}
The main failure mode for the vanilla heatmap network (conv1-conv8)
occurs when multiple modes are predicted and the wrong one is selected
(and the resulting poses are often kinematically impossible for a
human to perform). 
Examples of these failures are shown in Fig~\ref{fig:failures}.
The spatial fusion layers resolve these failures.

\subsection{Comparison to state of the art}

\paragraph{Training.} 
We investigated a number of strategies for training on these datasets including training from scratch (using only the training data provided with the dataset), or training on one (\ie BBC Pose) and fine-tuning on the others.  
We found that provided the first and last layers of the Spatial Net are initialized from (any) trained heatmap network, the rest can be trained either from scratch or fine-tuned with similar performance. 
We hypothesise this is because the datasets are very different -- BBC Pose contains long-sleeved persons, ChaLearn short-sleeved persons and Poses in the Wild contains non-frontal poses with unusual viewing angles.
For all the results reported here we train BBC Pose from scratch, initialize the first and last layer from this, and fine-tune on training data of other datasets.

\begin{figure*}
\vspace{-0.1cm}
\centering
\includegraphics[width=\linewidth]{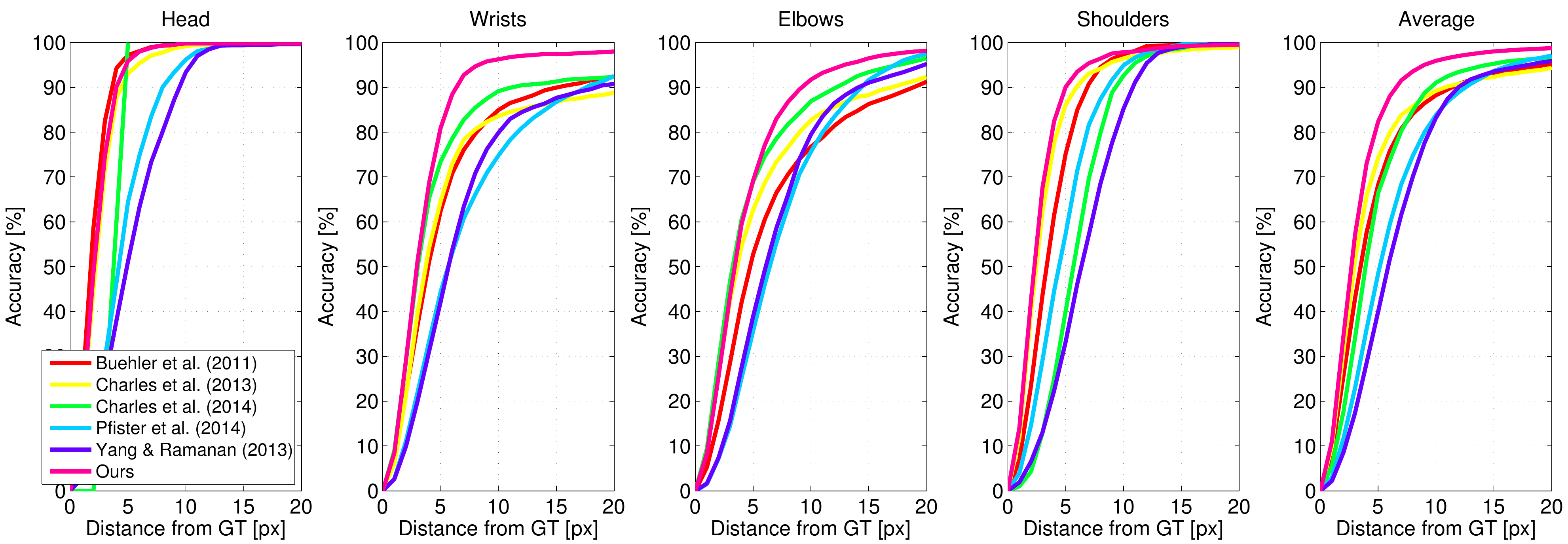}
\vspace{-0.4cm}
\caption{{\bf Comparison to the state of the art on BBC Pose.} 
Plots show accuracy per joint type (average over left and right body parts) as the allowed distance from manual ground truth is increased. 
We outperform all previous work by a large margin; notice particularly the performance for wrists, where we outperform the best competing method with an addition of 10\% at $d=6$.
Our method uses SpatialNet Flow Extended.
Pfister~\etal~(2014) uses Extended BBC Pose; Buehler~\etal, Charles~\etal and Yang~\& Ramanan use BBC Pose.
}
\label{fig:previouswork}
\end{figure*}

\vspace{-0.4cm}\paragraph{BBC Pose.}
Fig~\ref{fig:previouswork} shows a comparison to the state of the art on
the BBC Pose dataset.  We compare against all previous reported
results on the dataset.  These include Buehler~\etal~\cite{Buehler11},
whose pose estimator is based on a pictorial structure model;
Charles~\etal (2013)~\cite{Charles13a} who uses a Random Forest;
Charles~\etal (2014)~\cite{Charles14} who predict joints sequentially
with a Random Forest; Pfister~\etal (2014)~\cite{Pfister14a} who use a
deep network similar to our CoordinateNet (with multiple input frames); and 
the deformable part-based model of
Yang~\& Ramanan
(2013)~\cite{Yang13a}.

We outperform all previous work by a large margin, with a particularly noticeable gap for wrists (an addition of 10\% compared to the best competing method at $d=6$).

\begin{figure}[h]
\centering
\vspace{-0.1cm}
\includegraphics[width=\linewidth]{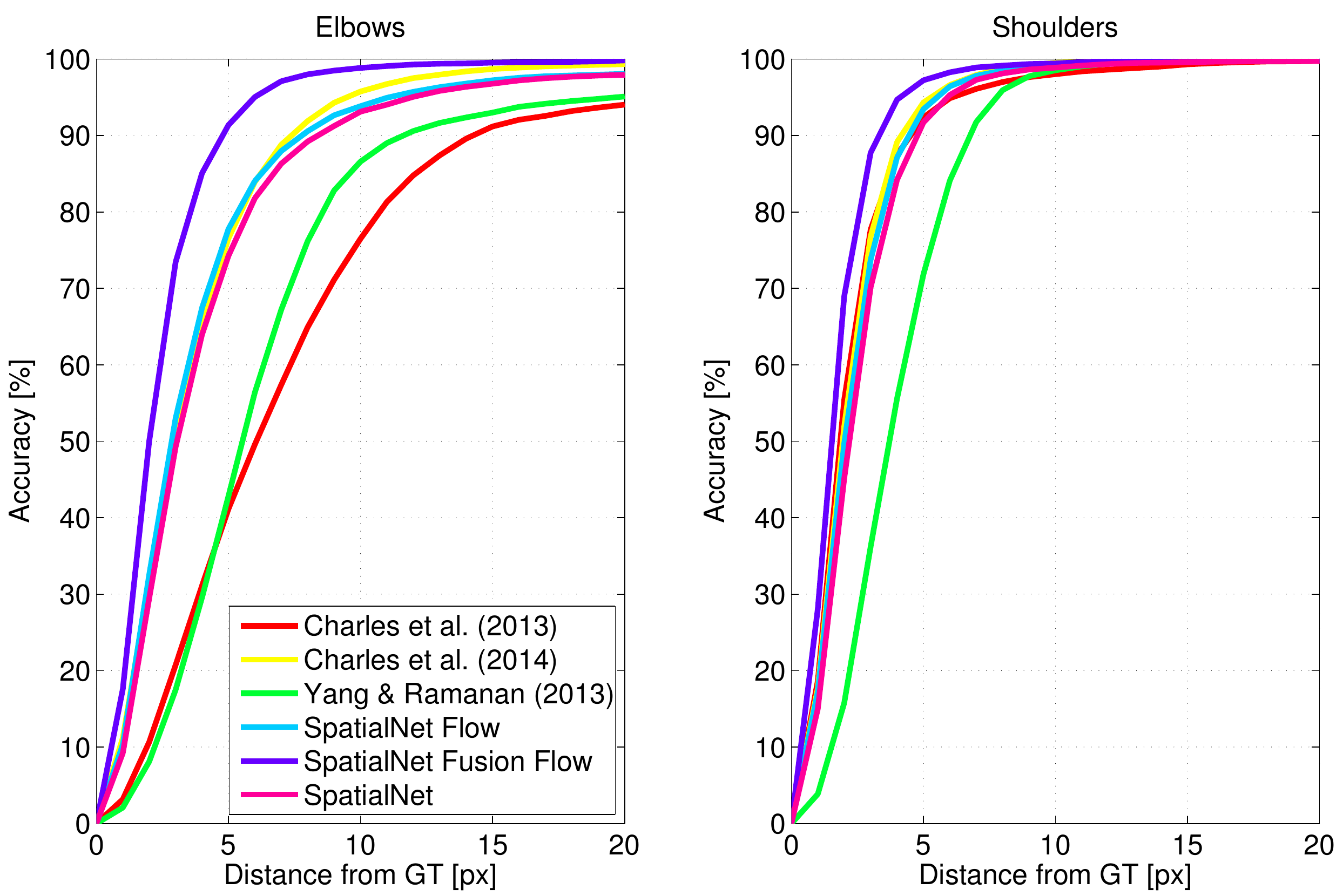}
\vspace{-0.4cm}
\caption{{\bf ChaLearn: elbows \& shoulders.}}
\label{fig:chalearn_2}
\end{figure}

\vspace{-0.4cm}\paragraph{Chalearn.}
Figs~\ref{fig:chalearn}~\&~\ref{fig:chalearn_2} show a comparison to the state of the art on ChaLearn.
We again outperform the state of the art even without optical flow (an improvement of 3.5\% at $d=6$), and observe a further boost by using optical flow (beating state of the art by an addition of 5.5\% at $d=6$), and a significant further improvement from using a deeper network (an additional 13\% at $d=6$).

\begin{figure}
\centering
\vspace{-0.2cm}
\includegraphics[width=\linewidth]{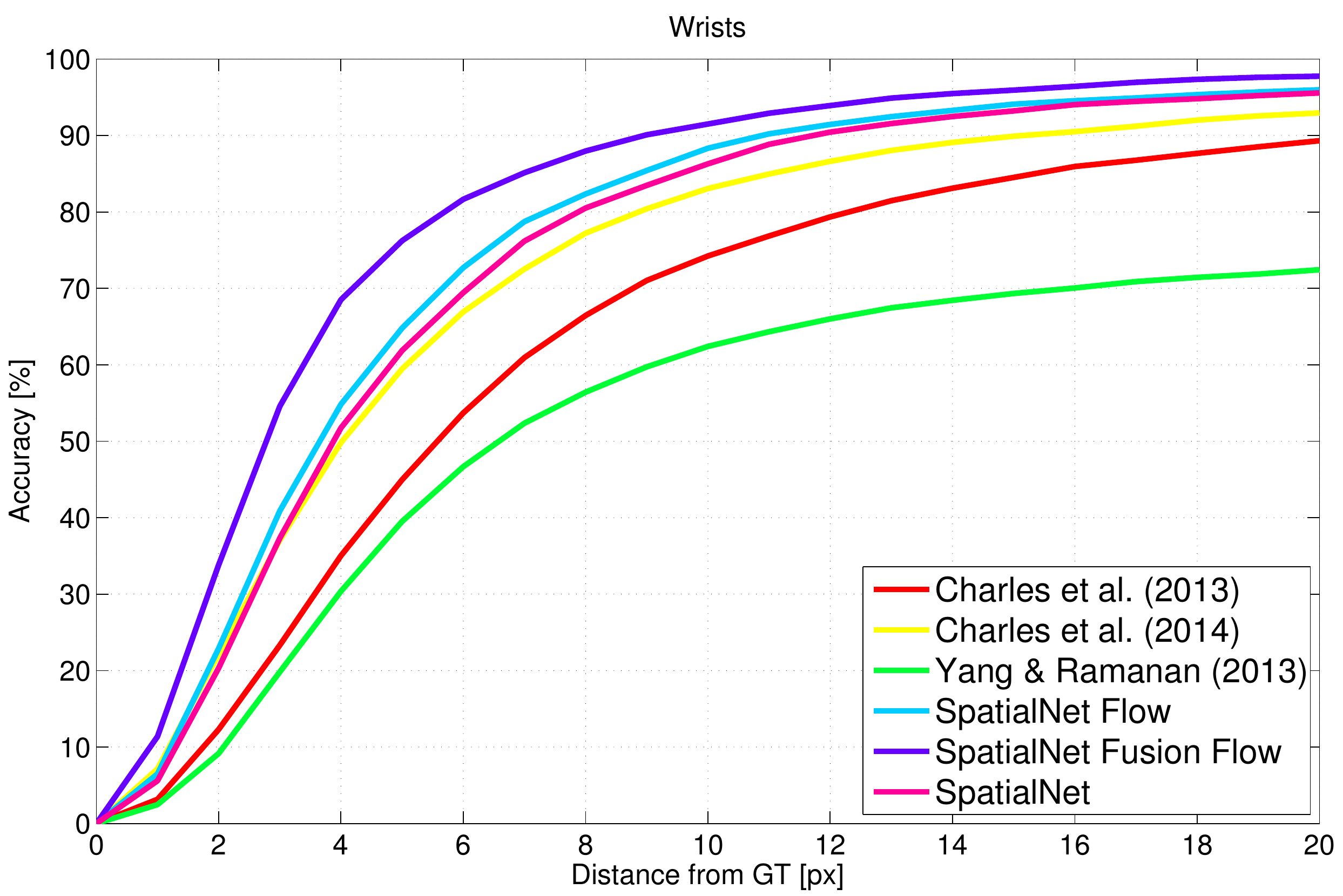}
\vspace{-0.4cm}
\caption{{\bf Comparison to the state of the art on ChaLearn.} 
Our method outperforms state of the art by a large margin (an addition of 19\% at $d=4$). }
\label{fig:chalearn}
\end{figure}

\begin{figure}
\centering
\vspace{-0.1cm}
\includegraphics[width=\linewidth]{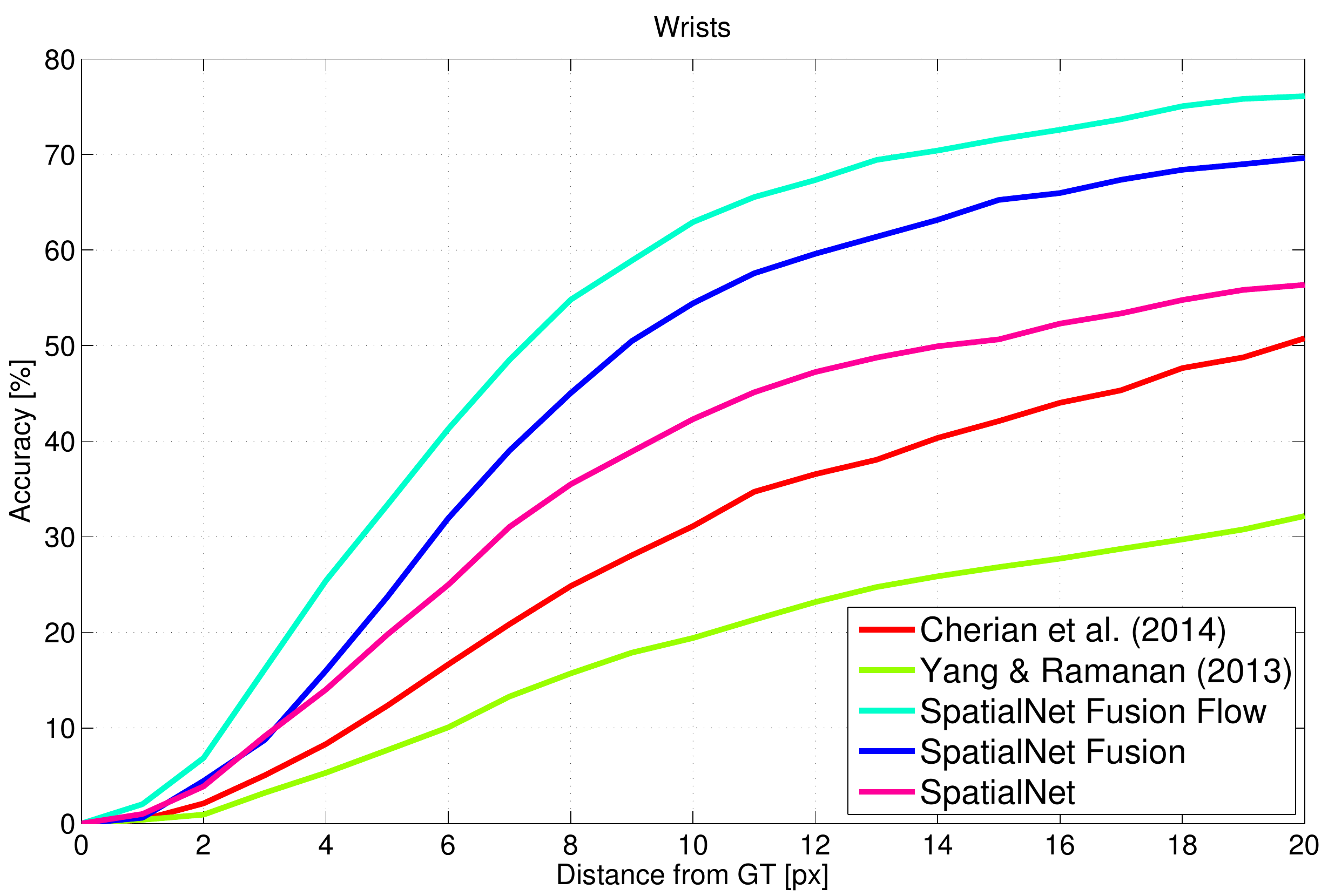}
\vspace{-0.4cm}
\caption{{\bf Comparison to state of the art on Poses in the Wild.} 
Our method outperforms state of the art by a large margin (an addition of 30\% at $d=8$, with 10\% from flow).
}
\label{fig:posesinthewild}
\end{figure}

\begin{figure}[h]
\centering
\vspace{-0.2cm}
\includegraphics[width=\linewidth]{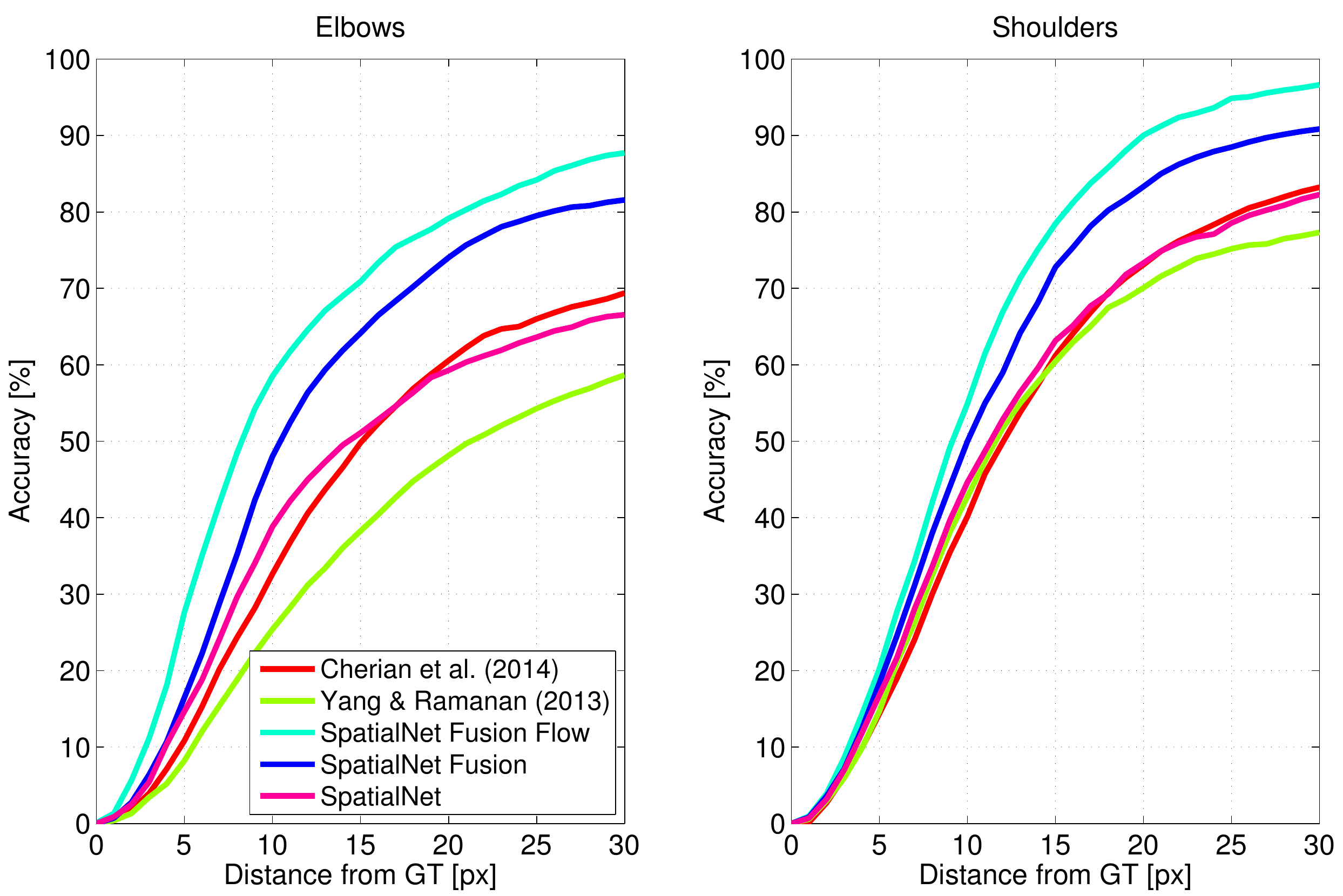}
\vspace{-0.4cm}
\caption{{\bf Poses in the Wild: elbows \& shoulders.} }
\label{fig:posesinthewild_2}
\end{figure}

\begin{figure*}
\vspace{0.4cm}
\centering
\includegraphics[width=\linewidth]{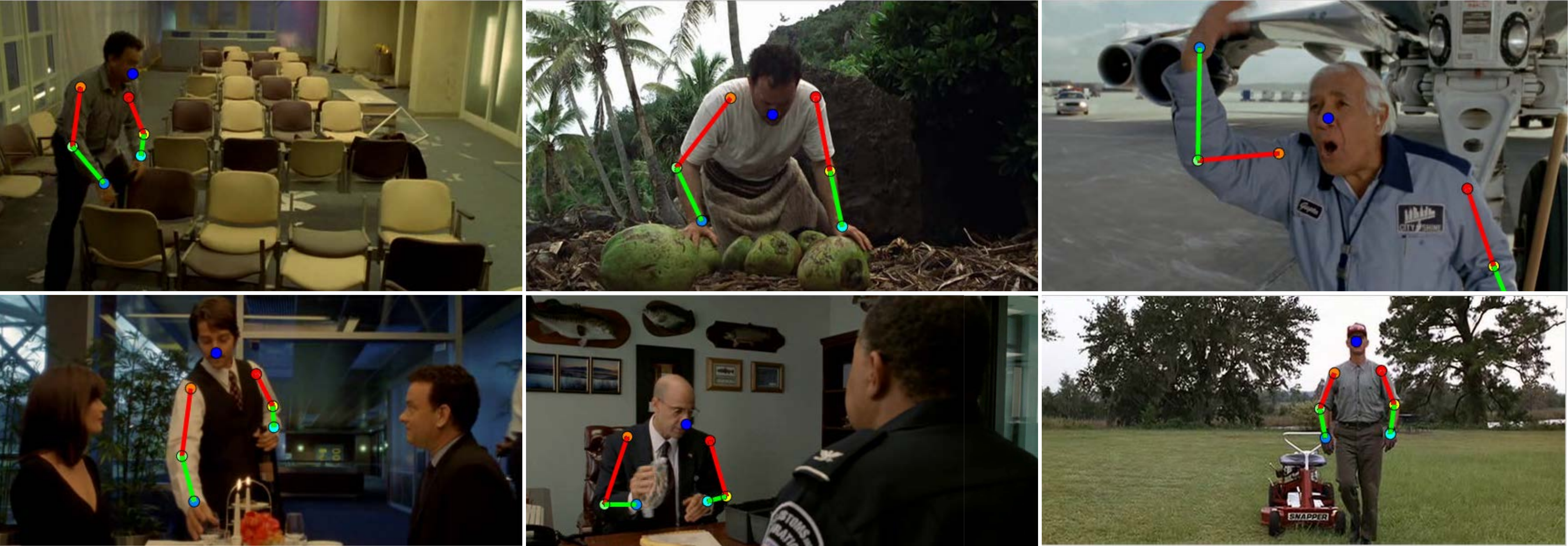}
\vspace{-0.2cm}
\caption{{\bf Example predictions on a variety of videos in Poses in the Wild.} }
\label{fig:posesinthewildexs}
\end{figure*}

\vspace{-0.4cm}\paragraph{Poses in the Wild.}
Figs~\ref{fig:posesinthewild}~\&~\ref{fig:posesinthewild_2} show a comparison to the state of the art on Poses in the Wild.
We replicate the results of the previous state of the art method using code provided by the authors~\cite{Cherian14}.
We outperform the state of the art on this dataset by a large margin (an addition of 30\% for wrists and 24\% for elbows at $d=8$).
Using optical flow yields a significant 10\% improvement for wrists and 13\% for elbows at $d=8$.
Fig~\ref{fig:posesinthewildexs} shows example predictions.

\begin{figure}
\centering
\vspace{-0.2cm}
\includegraphics[width=\linewidth]{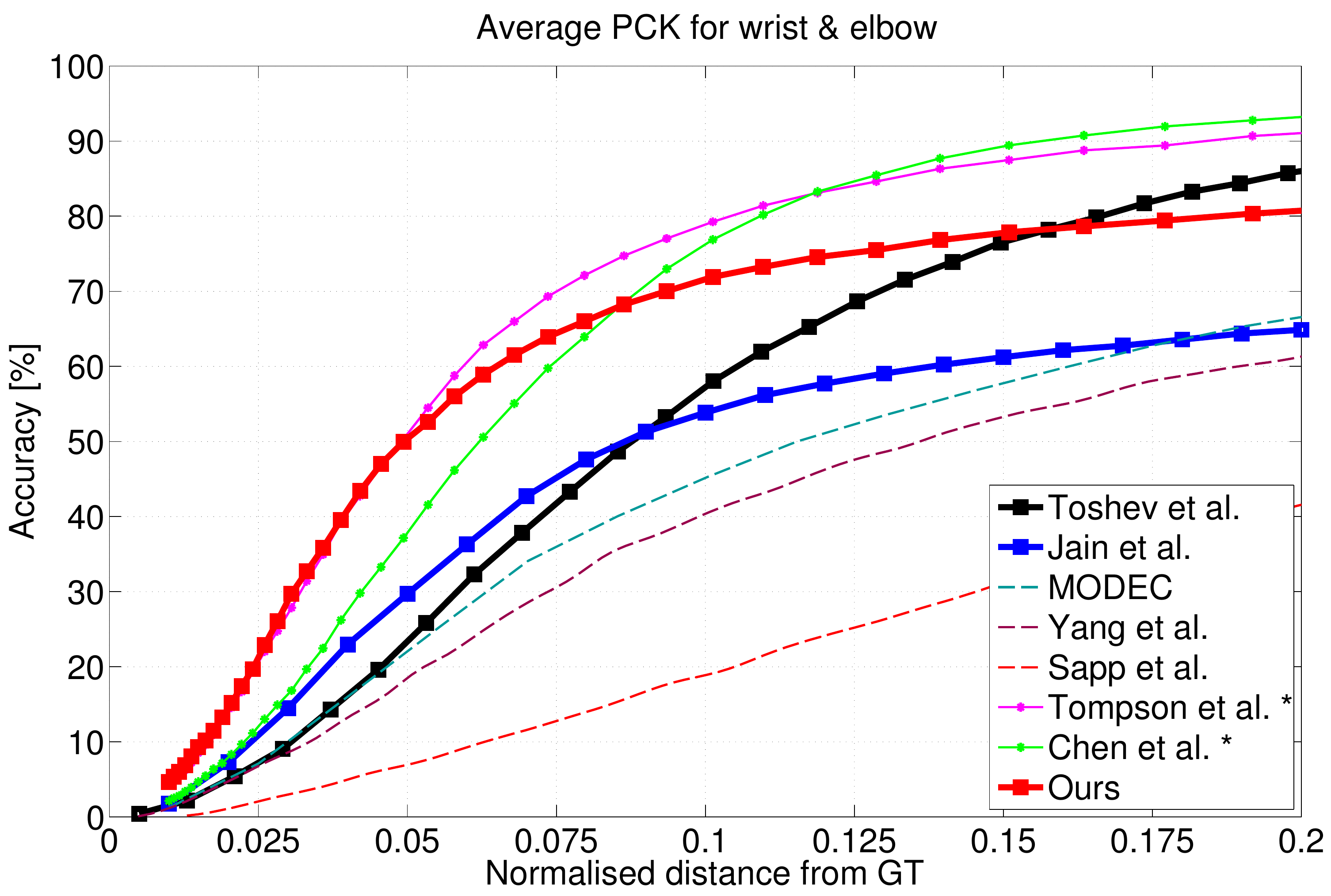}
\vspace{-0.4cm}
\caption{{\bf Comparison to state of the art on FLIC.}
Solid lines represent deep models; methods with a square ($\blacksquare$) are without a graphical model; methods with an asterisk (*) are with a graphical model.
Our method outperforms competing methods without a graphical model by a large margin in the high precision area (an addition of 20\% at $d=0.05$).
}
\vspace{0.2cm}
\label{fig:flic}
\end{figure}

\vspace{-0.4cm}\paragraph{FLIC.}
Fig~\ref{fig:flic} shows a comparison to the state of the art on FLIC.
We outperform all pose estimation methods that don't use a graphical model, and match or even slightly outperform graphical model-based methods~\cite{Chen14,Tompson14a} in the very high precision region ($<0.05$ from GT).
The increase in accuracy at $d=0.05$ is 20\% compared to methods not using a graphical model, and 12\% compared to~\cite{Chen14} who use a graphical model.
Tompson~\etal is~\cite{Tompson14a}; Jain~\etal is \cite{Jain14}.
Predictions are provided by the authors of~\cite{Chen14,Tompson14a} and evaluation code by the authors of~\cite{Tompson14a}.

\section{Conclusion}

We have presented a new architecture for pose estimation in videos
that is able to utilizes appearances across multiple frames. The
proposed ConvNet is a simple, direct method for regressing heatmaps,
and its performance is improved by combining it with optical flow and
spatial fusion layers.
We have also shown that our method
outperforms the state of the art on three large video pose estimation
datasets.
Further 
improvements may be obtained by using additional 
inputs for the spatial ConvNet,
for example multiple RGB frames~\cite{Pfister14a} or optical
flow~\cite{Jain14a} -- although prior work has shown little
benefit from this so far.

The benefits of aligning pose estimates from multiple frames using
optical flow, as presented here, are complementary to architectures
that explicitly add spatial MRF and refinement
layers~\cite{Tompson14,Tompson14a}.

Finally, we have demonstrated the architecture for human pose estimation, but 
a similar optical flow-mediated combination of information could be used
for other tasks in video, including classification and segmentation.

\paragraph{\bf Acknowledgements:} 
Financial support was provided by Osk.\ Huttunen Foundation and EPSRC grant EP/I012001/1.

\afterpage{\clearpage}

\begin{figure*}[t]
\centering
\includegraphics[width=\linewidth]{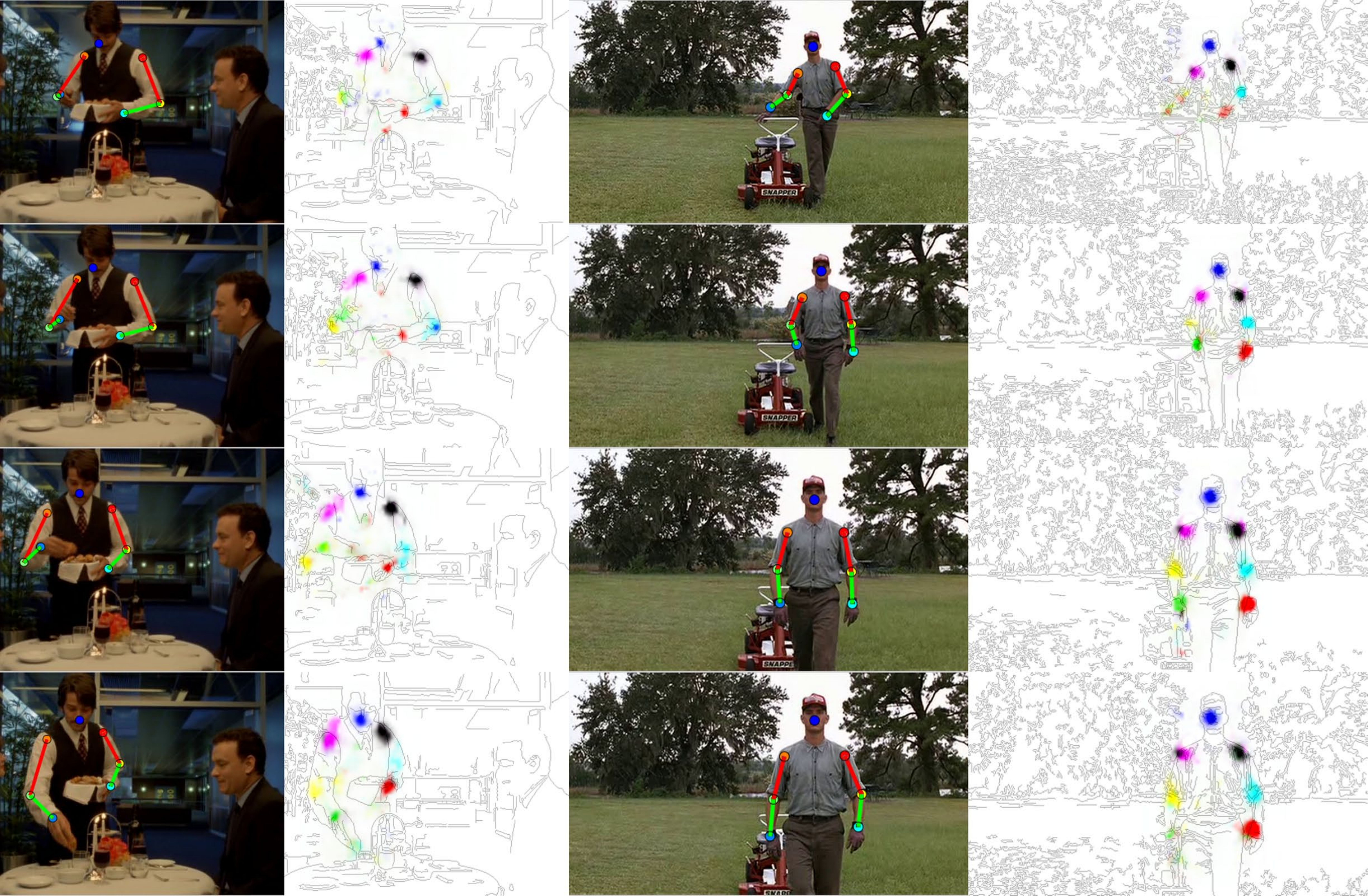}
\caption{{\bf Example predictions on two videos in Poses in the Wild.} 
Predictions and the corresponding heatmaps are shown.
}
\label{fig:piw_two}
\end{figure*}

\afterpage{\clearpage}

\begin{figure*}[p]
\centering
\includegraphics[width=\linewidth]{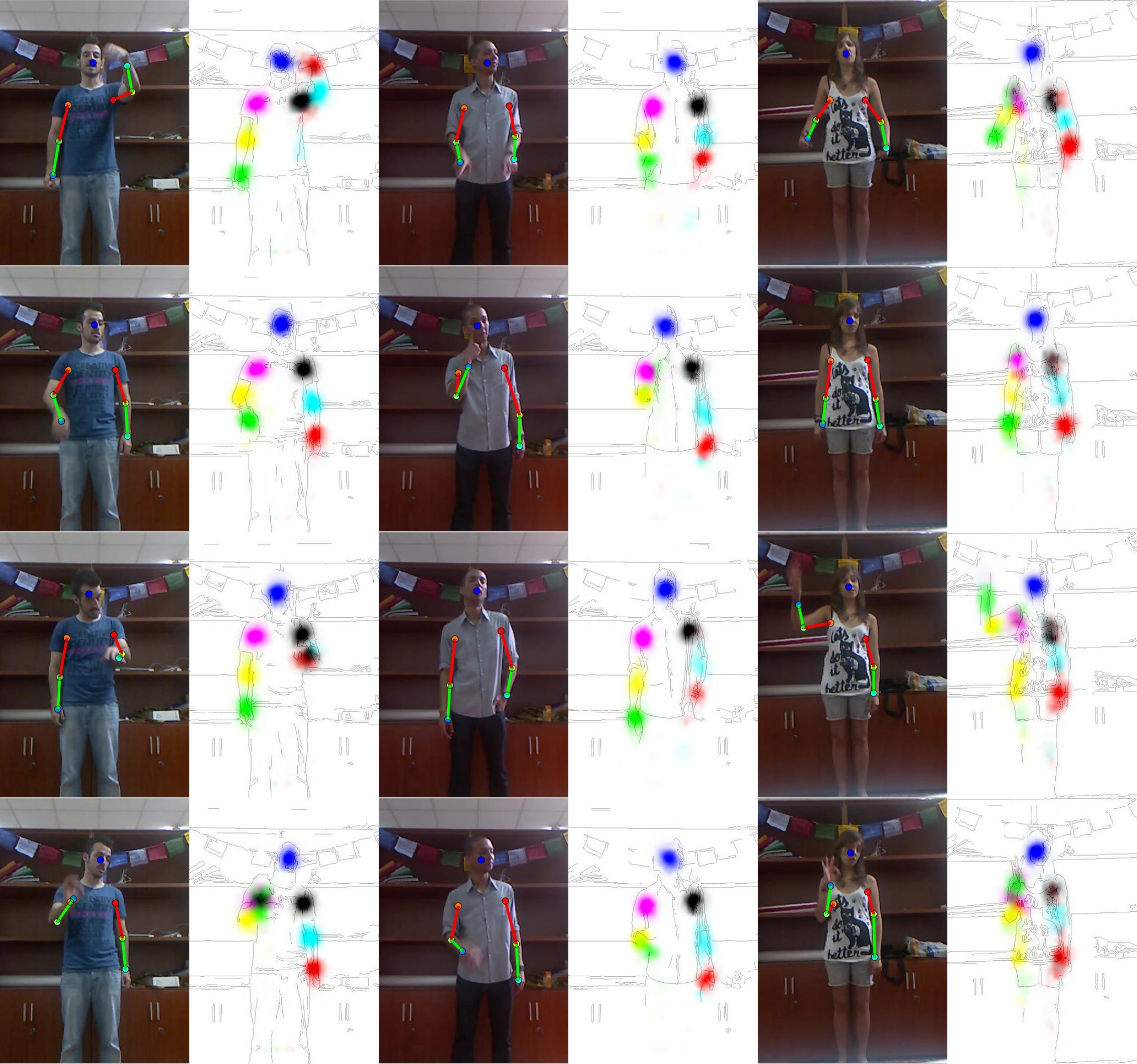}
\caption{{\bf Example predictions on ChaLearn.} 
}
\label{fig:chalearn}
\end{figure*}

\begin{figure*}[p]
\centering
\includegraphics[width=\linewidth]{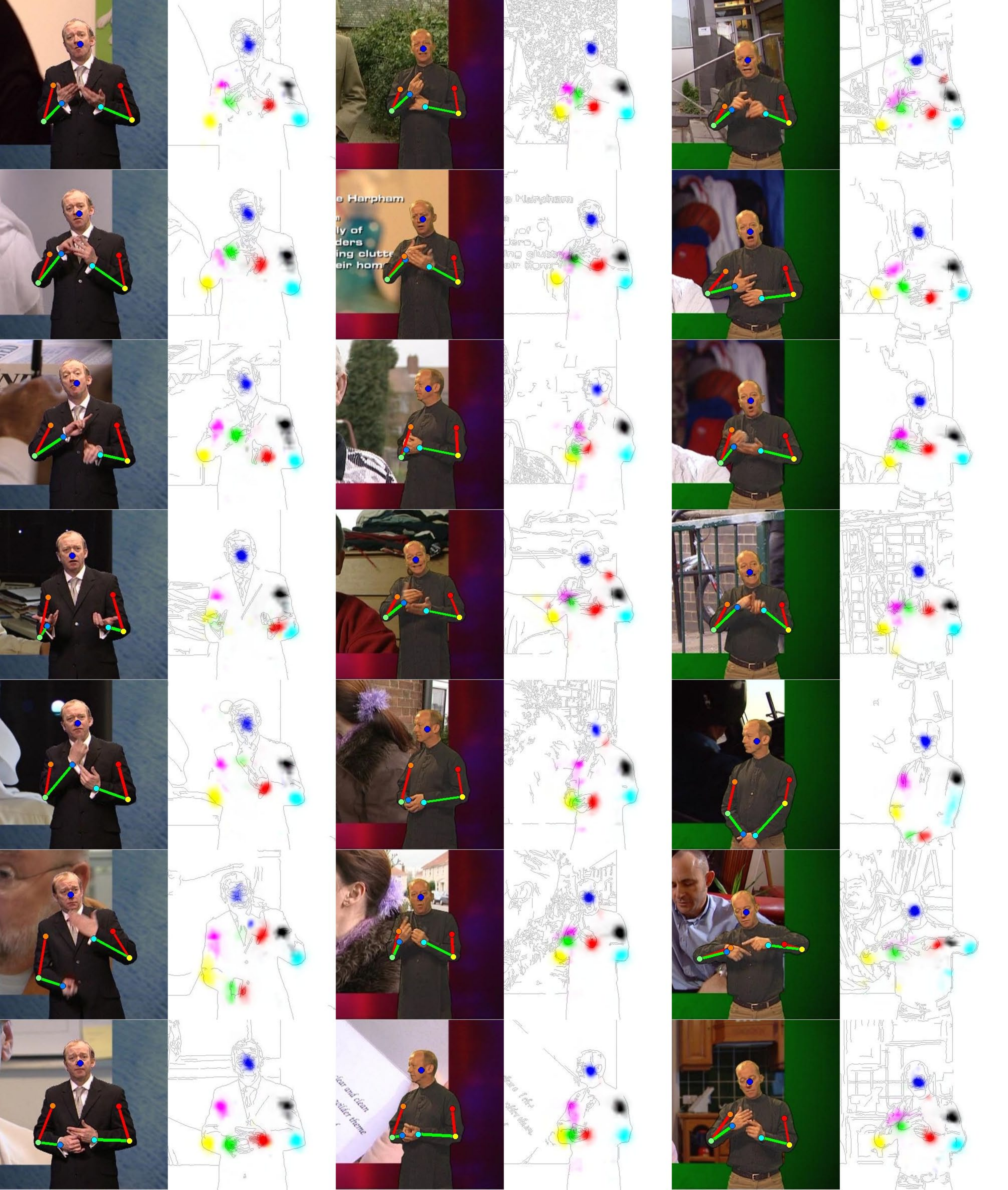}
\caption{{\bf Example predictions on BBC Pose.} 
}
\label{fig:bbc}
\end{figure*}

\newpage

{\small
\bibliographystyle{ieee}
\bibliography{bib/shortstrings,bib/vgg_local,bib/vgg_other,mybib}
}

\end{document}